%% file: main.tex
\documentclass[runningheads]{llncs}

\usepackage{eccv}

\usepackage{eccvabbrv}

\usepackage{graphicx}
\usepackage{booktabs}
\usepackage{tabularx}
\usepackage{amsfonts}
\usepackage{amsmath}
\usepackage{amssymb}
\usepackage{multirow}
\usepackage{float}
\usepackage{color}
\usepackage{colortbl}
\usepackage{xcolor}
\usepackage{adjustbox}
\usepackage{csquotes}
\usepackage{soul}
\usepackage{tabularray}
\usepackage{url}
\usepackage{wrapfig}
\usepackage[symbol]{footmisc}

\usepackage[accsupp]{axessibility}  % Improves PDF readability for those with disabilities.

\usepackage{hyperref}

\usepackage{orcidlink}

\definecolor{Gray}{gray}{0.96}

\begin{document}

\title{NeuSDFusion: A Spatial-Aware Generative Model for 3D Shape Completion, Reconstruction, and Generation} 

\titlerunning{NeuSDFusion: A Spatial-Aware Generative Model}

\renewcommand{\thefootnote}{\fnsymbol{footnote}}

\author{
Ruikai~Cui \inst{1} \thanks{The contribution of Ruikai~Cui, Han~Yan and Zhennan~Wu was made during an internship at Tencent XR Vision Labs.} \and
Weizhe~Liu \inst{2} \thanks{Corresponding author} \and
Weixuan~Sun \inst{2}\and
Senbo~Wang \inst{2}\and
Taizhang~Shang \inst{2}\and
Yang~Li \inst{2} \and
Xibin~Song \inst{2}\and
Han~Yan \inst{3}\and
Zhennan~Wu \inst{4}\and
Shenzhou~Chen \inst{2}\and
Hongdong~Li \inst{1} \and
Pan~Ji \inst{2}
}
\authorrunning{R.~Cui et al.}

% \institute{Australian National University \and
% Tencent XR Vision Labs \and
% Shanghai Jiao Tong University \and
% The University of Tokyo
% }
\institute{
\mbox{%
\inst{1}Australian National University \quad
\inst{2}Tencent XR Vision Labs
}
\mbox{
\inst{3}Shanghai Jiao Tong University \quad
\inst{4}The University of Tokyo
}
}
\maketitle

\begin{abstract}
3D shape generation aims to produce innovative 3D content adhering to specific conditions and constraints. Existing methods often decompose 3D shapes into a sequence of localized components, treating each element in isolation without considering spatial consistency. As a result, these approaches exhibit limited versatility in 3D data representation and shape generation, hindering their ability to generate highly diverse 3D shapes that comply with the specified constraints. In this paper, we introduce a novel spatial-aware 3D shape generation framework that leverages 2D plane representations for enhanced 3D shape modeling. To ensure spatial coherence and reduce memory usage, we incorporate a hybrid shape representation technique that directly learns a continuous signed distance field representation of the 3D shape using orthogonal 2D planes. Additionally, we meticulously enforce spatial correspondences across distinct planes using a transformer-based autoencoder structure, promoting the preservation of spatial relationships in the generated 3D shapes. This yields an algorithm that consistently outperforms state-of-the-art 3D shape generation methods on various tasks, including unconditional shape generation, multi-modal shape completion, single-view reconstruction, and text-to-shape synthesis. Our project page is available at \url{https://weizheliu.github.io/NeuSDFusion/}.

\end{abstract}

\section{Introduction}
\label{sec:intro}

\begin{figure}[t]
    \centering
    \includegraphics[width=\linewidth]{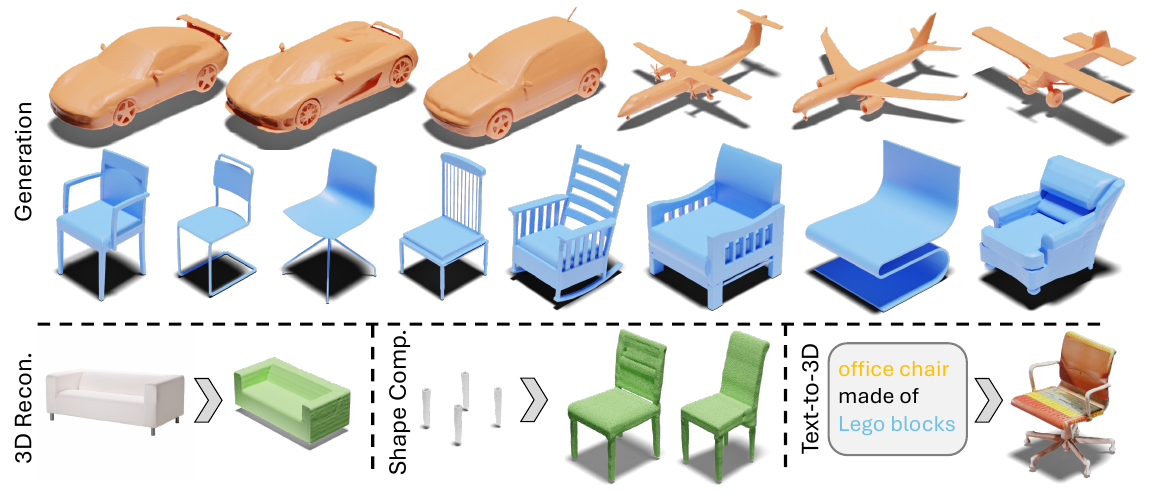}
    \caption{\textbf{NeuSDFusion} demonstrates exceptional performance in generating high-quality, diverse shapes with smooth surfaces and detailed structures, showcasing its capability in 3D shape synthesis across various tasks including unconditional generation, single-view reconstruction, shape completion, and text-to-3D synthesis.}
    \label{fig:intro}
\end{figure}

% Opening Statement on the Importance of 3D Shape Generation:
In the fast-paced realm of computer vision and graphics, generative modeling has emerged as a crucial aspect of 3D shape creation, propelling advancements in 3D content generation. Recently, the remarkable achievements of generative models in 2D image synthesis and video production have inspired significant interest in exploring generative techniques, such as diffusion models~\cite{rombach2022high,ramesh2022hierarchical}, in the creation of 3D content. This development reflects a broader transition towards more sophisticated and flexible 3D shape generation methods, fueled by the growing demand for high-quality 3D content across diverse industry applications. 

Recent studies~\cite{hong2023lrm,gao2022get3d,poole2022dreamfusion} have focused on understanding 3D geometry by leveraging multi-view supervision with predefined camera positions and incorporating inductive biases through neural rendering~\cite{mildenhall2021nerf} techniques. While these approaches yield impressive results, the training process can be time-consuming and often overlooks accessible 3D data that could be harnessed to derive more effective shape priors. To directly exploit such shape priors from 3D data, contemporary methods~\cite{cheng2023sdfusion, li2023dqd} typically model the distribution of 3D objects using explicit 3D representations. These techniques frequently employ Truncated Signed Distance Function (T-SDF) as the 3D object representation and utilize an encoder to embed an object directly into a latent representation. However, the T-SDF representation usually employs a 3D voxel representation to store the distance field, which is generally memory-intensive, hindering the capture of detailed shape characteristics due to memory constraints. 
% Furthermore, T-SDF truncates distance values, leading to a discontinuous representation that may not be optimal for applications requiring smooth surfaces or continuous fields. 
Inspired by recent efforts~\cite{chan2022efficient} in using 2D plane representations for 3D shape modeling, some works~\cite{chou2023diffusion,gupta20233dgen} opt to encode an object as a tri-plane representation for embedding and employ point clouds as an intermediate representation. However, such an intermediate representation necessitates extra encoding networks, resulting in reduced efficiency. NFD~\cite{shue20233d}, a work closely related to ours, utilizes tri-planes as the 3D object representation. However, their tri-planes are learned from occupancy grids, so that the representation capability is constrained by the grid resolution. This limitation highlights the need for more advanced 3D shape representations that can effectively capture spatial coherence and generate high-quality 3D shapes, paving the way for further progress in the field of 3D generative models.

Furthermore, while the tri-plane representation has been proposed to preserve 3D information of objects by utilizing three orthogonal 2D planes, existing approaches often overlook the 3D correlations among these planes. Previous works~\cite{shue20233d,chou2023diffusion} concatenate the three planes in the channel dimension and treat them as RGB images, even though there is no explicit relationship between the same coordinates on different planes. These methods neglect the synergy between planes, leading to disorder in the feature learning process, inferior shape details, and artifacts. An attempt to address cross-plane correlation is the 3D-aware convolution proposed by Rodin~\cite{wang2023rodin}. However, this method relies on an approximated feature aggregation through a pooling operation to tackle the computational cost, which results in displaced shape details and hampers in-plane communication.

To this end, we present NeuSDFusion, a framework for generating high-fidelity 3D objects, as depicted in Fig.~\ref{fig:intro}. Unlike previous methods that either rely on rendering-based techniques, ignoring available explicit 3D priors, or employ  raw 3D data (point clouds~\cite{vahdat2022lion}, mesh~\cite{liu2022meshdiffusion}, occupancy grid~\cite{shue20233d,peng2020convolutional}, or SDF~\cite{mittal2022autosdf,li2023dqd}) as 3D representations, limited by memory constraints, we introduce a hybrid tri-plane SDF representation named NeuSDF. This representation expressively embeds 3D objects as orthogonal 2D planes, preserving 3D structural information and generating smooth surfaces, while also being computationally efficient and capable of handling high-resolution details.

Our proposed approach consists of a three-stage pipeline. In the first stage, we fit each object in a dataset with a NeuSDF representation. To achieve generative modeling of this representation and retain the topological structure, we propose a specially designed autoencoder in the second stage. This autoencoder compresses the raw NeuSDF representation into a compact latent representation that preserves spatial correspondence among different planes. In the final stage, we adapt a diffusion model to synthesize these compressed latent representations of objects with various conditioning signals. The generated latent representations can then be further decoded into a new NeuSDF representation and ultimately transformed into a novel 3D object via marching cubes shape extraction~\cite{lorensen1998marching}.

We demonstrate the capability of our framework in various settings, including unconditional generation, multi-modal shape completion, single-view reconstruction, and text-guided generation. In summary, our contributions are: 
\begin{itemize} 
    \item A novel pipeline capable of generating high-fidelity 3D shapes under various conditions. 
    \item A hybrid 3D representation that captures highly detailed surface shapes with minimal memory consumption. 
    \item A spatial-aware autoencoder structure designed to maintain spatial coherence for our proposed hybrid 3D representation. 
    \item Extensive experiments showing that our method achieves state-of-the-art performance on various 3D shape generation benchmarks, surpassing previous methods in terms of both generation quality and diversity. 
\end{itemize}

\section{Related Works}
\label{sec:related}

\subsection{Generative Modeling of 3D Shapes}
\label{sec:generation}
In recent years, numerous works have been proposed to generate 3D shapes, with existing 3D generative models built on various frameworks. This includes generative adversarial networks (GANs)\cite{wu2016learning,achlioptas2018learning,zheng2022sdfstylegan}, variational autoencoders (VAEs)\cite{tan2018variational,gao2021tm,kim2021setvae, mittal2022autosdf}, normalizing flows~\cite{yang2019pointflow}, autoregressive models~\cite{zhang20223dilg}, energy-based models~\cite{xie2021generative,cui2022energy}, and more recently, denoising diffusion probabilistic models (DDPMs)~\cite{zhou2021pvd,wu2024blockfusion,chen2023single,erkocc2023hyperdiffusion,hui2022neural}. Luo \etal\cite{luo2021diffusion} pioneered the application of DDPMs for modeling raw point clouds. Building on the success of latent diffusion models~\cite{rombach2022high} in 2D image generation, several works~\cite{vahdat2022lion,li2023dqd,cheng2023sdfusion} explored generative modeling of 3D shapes in latent space to reduce computational complexity and enhance generation quality. Within this context, LION~\cite{vahdat2022lion} utilized point clouds to represent 3D objects, while 3DQD~\cite{li2023dqd}, SDFusion~\cite{cheng2023sdfusion}, and DiffusionSDF~\cite{chou2023diffusion} employed the Signed Distance Function (SDF) for 3D shape representation. Concurrently, other research efforts have investigated DDPMs for 3D shape generation with varied representations, such as mesh~\cite{liu2022meshdiffusion}, occupancy grid~\cite{shue20233d}, and neural radiance fields~\cite{muller2023diffrf}. 
Despite these advancements, a common limitation among most approaches is their inadvertent neglect of spatial coherence during the generation process. Addressing this gap, a recent study by Wang \etal~\cite{wang2023rodin} introduces a method that maintains spatial correspondence through the use of 3D-aware convolution on rolled-out tri-planes. Nevertheless, this technique employs axis-wise pooling to downsample plane features to a vector and only aggregates such downsampled features without enforcing in-plane communication. As a result, it lacks context information and fails to generate smooth shape surfaces. This highlights the need for more advanced methods that can effectively capture spatial coherence and generate high-quality 3D shapes.

\subsection{Representation of 3D Shapes}

% These methods typically model 3D objects as three axis-aligned planes~\cite{chan2022efficient} and use convolutional neural networks to generate such 2D planes. 
Various 3D shape representations have been investigated to facilitate the 3D synthesis task. We categorize them into two primary classes: rendering-based and rendering-free methods. Rendering-based approaches~\cite{karnewar2023holodiffusion, hong2023lrm}, such as LRM~\cite{hong2023lrm}, Zero123~\cite{liu2023zero}, employ multi-view images to learn object geometry via volume rendering~\cite{mildenhall2021nerf}. However, these methods overlook the available 3D objects as priors, leading to coarse shapes that often lack fine geometry details. On the other hand, rendering-free methods model the distribution of 3D objects using raw 3D representations, such as point clouds~\cite{cui2024lam3d}, mesh~\cite{alliegro2023polydiff}, binary occupancy~\cite{yan2022shapeformer,zhang20233dshape2vecset}, and raw SDF~\cite{zheng2023locally,zhang2024functional,yariv2024mosaic}. These methods necessitate careful network design (\eg, treating point clouds as sets~\cite{luo2021diffusion} or handling mesh edges~\cite{liu2022meshdiffusion}), and may struggle to represent intricate object structures. SDFusion~\cite{cheng2023sdfusion} and 3DQD~\cite{li2023dqd} employ T-SDF to represent 3D objects. However, direct use 3D representation requires substantial memory, while most of the 3D space remains devoid of shape surfaces. Consequently, these 3D representation approaches are generally less effective compared to 2D plane formulations~\cite{chan2022efficient}. NDF~\cite{shue20233d} was the first to introduce a tri-plane representation for modeling raw 3D data. However, their tri-plane representation was derived from a discrete occupancy grid, constraining shape details by the resolution of planes.

\section{Methodology}
\label{sec:method}

\begin{figure}[t]
    \centering
    \includegraphics[width=\linewidth]{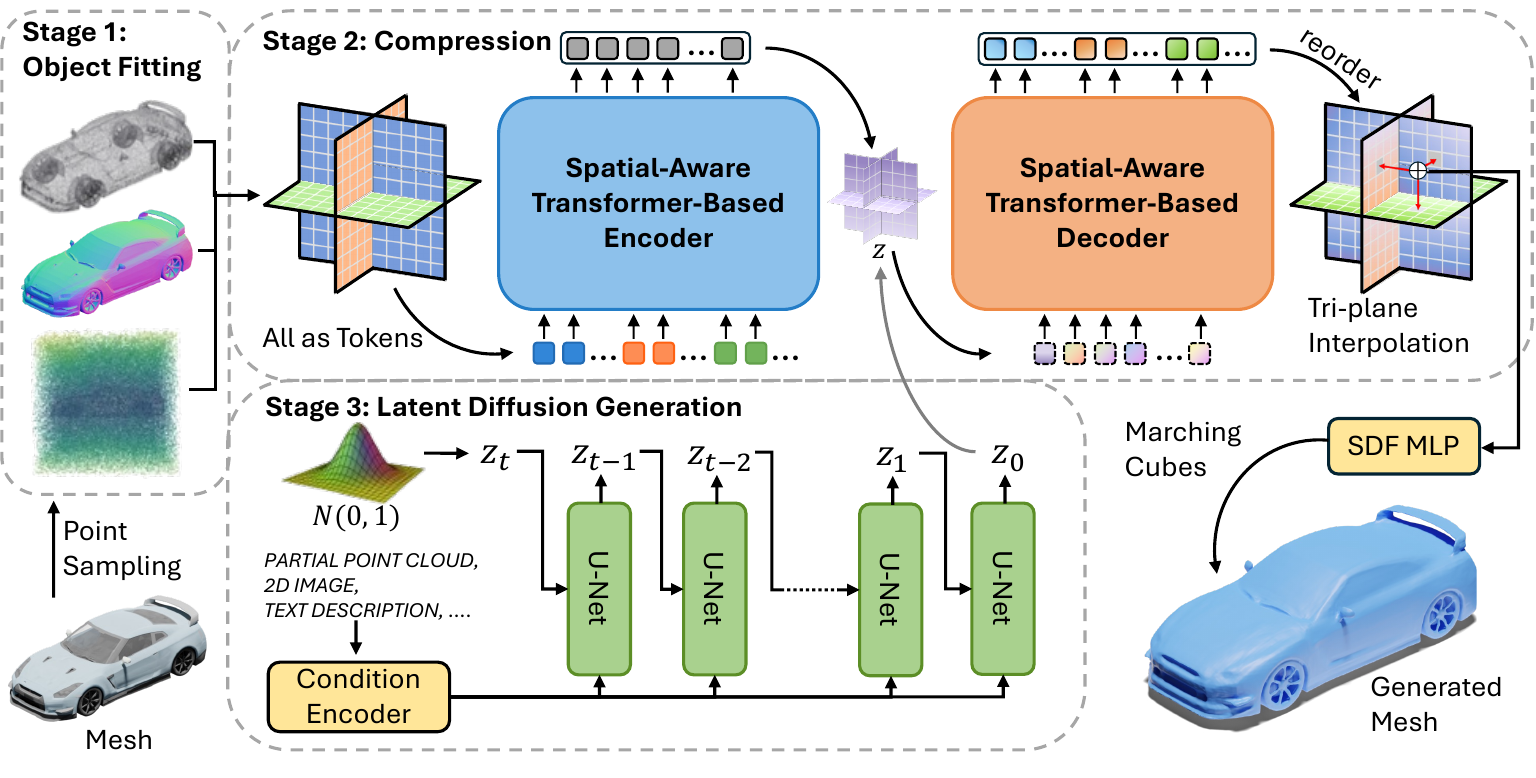}
    \caption{Our method follows a pipeline consisting of three stages. Given a raw mesh, we first sample surface points and space-filling points to adapt each mesh to a NeuSDF representation. In the second stage, we compress the raw tri-plane representation into latent tri-planes $z$ with a spatial-aware autoencoder. In the third stage, we train a latent diffusion model capable of generating tri-plane latent $z_0$ from a standard Gaussian under flexible conditions. During the inference phase, we input the generated latent $z_0$ into the decoder, and generate a mesh using the Marching Cubes algorithm by querying the signed distance value of any position via interpolating the reconstructed tri-plane.}
    \label{fig:pipeline}
\end{figure}

Our pipeline comprises three stages, as illustrated in Fig.~\ref{fig:pipeline}. Initially, we model each 3D object with a novel hybrid representation named NeuSDF, mapping the signed distance function into three orthogonal 2D planes (Sec.~\ref{sec:tri-plane}). 

In the second stage, we compress this 2D plane representation into latent features. To this end, we propose a transformer-based autoencoder, which efficiently encodes the tri-planes into a compact representation while preserving the 3D correlations between planes (Sec.~\ref{sec:ae}). In the final stage, we train a diffusion model capable of performing unconditional or guided generation with prompts from various modalities, including images, text, and point clouds (Sec.~\ref{sec:ldm}).

\subsection{Representing objects as NeuSDF}
\label{sec:tri-plane}

\smallskip
\noindent
\textbf{Formulation.} 
A signed distance function is a function that maps a spatial coordinate $(x, y, z)$ to a real value $s$, indicating the distance of the spatial position to the object surface. This function is commonly represented with a network or a vector representation~\cite{park2019deepsdf,mittal2022autosdf}. Instead, inspired by the recent success of the tri-plane representation~\cite{chan2022efficient}, we represent the geometry of a 3D shape using three axis-aligned planes, \ie the XY, YZ, and XZ planes. We then use a multi-layer perceptron (MLP)  to decode the tri-plane into signed distance values. Formally, we query the signed distance value of any 3D position $p \in \mathbb{R}^3$ by projecting it onto each of the three feature planes, retrieving the corresponding feature vector $F_{xy}, F_{xz}, F_{yz}$ via bilinear interpolation and aggregating the three feature vectors to obtain the feature vector for the query position. Then, we use an MLP $\phi$ to interpret the aggregated feature as a real value. The signed distance of any query position $p$ is acquired as follows:
\begin{equation}
    \Phi(p) = \text{MLP}_{\phi}(F_{xy} \oplus F_{xz}\oplus F_{yz}),
\end{equation}
where $\oplus$ denotes element-wise summation.

\smallskip
\noindent
\textbf{Point Sampling.} 
To effectively represent a 3D object as a NeuSDF, we sample SDF values to train the tri-planes. Given a 3D object mesh, we first preprocess it to transform it to a watertight shape using Blender’s \textit{voxel remeshing} tool. Then, we normalize the shape within the $[-1, 1]^3$ box. Subsequently, we randomly sample on-surface points $\Omega_0$ from the shape surface and uniformly sample off-surface points $\Omega$ with ground truth SDF values in the $[-1, 1]^3$ space. 
In addition to the signed distance supervision, we leverage normal direction as extra guidance to achieve detailed surface modeling. Consequently, we also sample the normal vector for each on-surface point. 

\smallskip
\noindent
\textbf{Fitting.} 
We employ an optimization-based approach to convert each 3D object into tri-planes, which will then be used for training our generative model. There are two learnable competent, \ie, the tri-planes and the MLP with parameters $\phi$. We jointly optimize both using the following geometry loss:
\begin{equation}\label{eq:L_geo}
    \mathcal{L}_{geo} = \mathcal{L}_{sdf} + \mathcal{L}_{normal} + \mathcal{L}_{eikonal}.
\end{equation}
These three terms are defined as:
\begin{align}
    \mathcal{L}_{sdf} & = \lambda_1 \sum_{p \in \Omega_0} || \Phi(p) || + \lambda_2 \sum_{p \in \Omega} || \Phi(p) - d_p ||, \\
    \mathcal{L}_{normal} & = \lambda_3 \sum_{p \in \Omega_0} || \nabla_p \Phi(p) - n_p ||, \\
    \mathcal{L}_{eikonal} & = \lambda_4 \sum_{p \in \Omega_0} || ||\nabla_p \Phi(p)|| - 1 ||,
\end{align}
where $d_p$ and $n_p$ are ground truth SDF value and surface normal vector, respectively. 
The gradient 
$\nabla_p \Phi(p) = \left( 
\frac{\partial \Phi(p)}{\partial X}, 
\frac{\partial \Phi(p)}{\partial Y}, 
\frac{\partial \Phi(p)}{\partial Z} \right)$
represents the direction of the steepest change in SDF. It can be computed using finite differences, \eg, the partial derivative for the X-axis component
reads
$
\frac{\partial \Phi(p)}{\partial X} = 
\frac{\Phi(p + (\delta, 0, 0)) - \Phi(p - (\delta, 0, 0))}{2\delta}
$
where $\delta$ is the step size. The Eikonal loss constrains $\|\nabla_p \Phi(p)\|$ to
be $1$ almost everywhere, thus maintaining the intrinsic physical property of the signed distance function.

Our NeuSDF formulation learns continuous SDF values using a compact 2D plane representation, which stands in contrast to previous work. Existing methods either directly leverage T-SDF representation~\cite{cheng2023sdfusion,li2023dqd}, which is memory-intensive and sensitive to noise, or utilize additional point cloud encoding networks~\cite{chou2023diffusion}, resulting in reduced efficiency. 
 
\subsection{Compression with Spatial-Aware Autoencoder}
\label{sec:ae}

Given that we individually embed each object as a tri-plane, directly training a diffusion model on these tri-planes would lead to two significant drawbacks: 1) an excessive number of parameters to diffuse, resulting in the model's inability to generalize and synthesize novel tri-planes, and 2) substantial model complexity, necessitating vast computational resources~\cite{rombach2022high}. To address these limitations, we introduce a spatial-aware autoencoder to compress tri-planes into latent representations. Specifically, given a tri-plane $x \in \mathbb{R}^{3 \times C \times H \times W}$, the encoder $\mathcal{E}$ encodes $x$ into a latent representation $z = \mathcal{E}(x)$, and the decoder $\mathcal{D}$ reconstructs the tri-plane from this latent representation. Thus, we have $\tilde{x} = \mathcal{D}(z) = \mathcal{D}(\mathcal{E}(x))$, where $z\in \mathbb{R}^{3 \times c \times h \times w}$, and $c$ is the feature dimension.

\begin{figure}[t]
    \centering
    \includegraphics[width=0.95\linewidth]{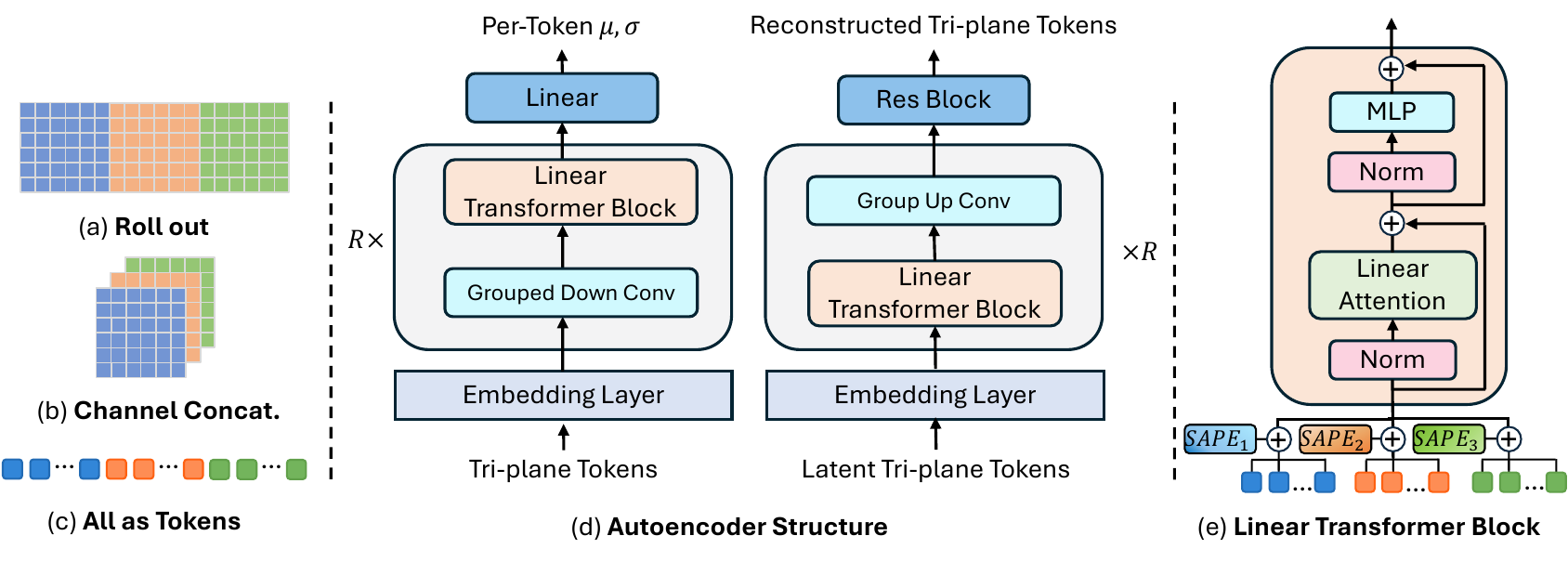}
    \caption{An illustration of the spatial-aware autoencoder design. Both (a) a roll-out mechanism and (b) a channel-wise concatenation strategy utilize a convolutional neural network to manipulate a 2D feature map, which leads to a contextual disorder. To address this issue, we propose the (c) \textit{all as tokens} operation which is designed to preserve spatial coherence. This operation is facilitated by (d) a transformer-based autoencoder structure and the implementation of (e) a spatial-aware position embedding (SAPE) technique.}
    \label{fig:autoencoder}
\end{figure}

\smallskip
\noindent
\textbf{Model Structure.} 
A tri-plane data structure differs from a grid-like image that can be efficiently processed via convolutional neural networks, such as the autoencoder used in Stable Diffusion~\cite{rombach2022high}. We aim to model and compress the global 3D information of tri-planes into compact latent representations while preserving the 3D correlations between planes. However, most existing methods overlook these correlations. For instance, \textit{roll out} used in Rodin~\cite{wang2023rodin}, and ~\cite{shue20233d,chou2023diffusion} naively concatenate tri-planes either horizontally or in the channel dimension to form a 2D feature map. Such approaches result in disorder in the feature learning process and generate artifacts. Moreover, Wang~\etal~\cite{wang2023rodin} attempts to address cross-plane correlation using 3D-aware convolution, but it relies on a feature aggregation using a pooling operation to tackle the computational cost and generate inferior results in our experiments. In summary, existing tri-plane autoencoders, which all adopt convolution, do not guarantee meaningful spatial correspondence among different planes. 

To effectively compress the tri-planes, an efficient and expressive structure capable of handling tri-plane features while maintaining spatial consistency is essential. We propose a transformer-based autoencoder designed to process and progressively downsample tri-plane features in accordance with their 3D relationship, as depicted in Fig.~\ref{fig:autoencoder}.

Our transformer-based autoencoder adheres to a U-shaped encoder-decoder structure. 
Both the encoder and decoder comprise several stages, where the downsampling operation and cross-plane correlation are separately handled within each stage. This design ensures a balanced approach to feature compression and spatial comprehension.
Within each stage, tri-plane features are first concatenated in the channel dimension. 
Subsequently, we use a group convolution to downsample each plane individually with a separate kernel,
allowing for more parameter-efficient compression while preserving the spatial relationships among the tri-plane features.
Following this, the tri-plane is flattened into a 1D sequence of tokens $x \in \mathbb{R}^{C \times 3HW}$ and input into a transformer block. 
This transformer block attends to different parts of the tri-plane representations and learn global relationships between three planes. 
However, a prominent limitation of the transformer structure is the computational overhead of the self-attention module, as its complexity grows quadratically with respect to the sequence length ($3 \times 64 \times 64 = 12288$ in the first stage in our setting).
This complexity poses a challenge to the scalability of the transformer-based model to handle tri-planes.
For example, LRM~\cite{hong2023lrm} only processes tri-planes with a resolution of $32$, leading to unsatisfactory details.
To overcome this difficulty, we utilize the linear attention mechanism proposed in ~\cite{qin2022devil, qin2023scaling}. 
Linear attention considerably reduces computational complexity, enabling us to directly operate on high-resolution tri-planes and achieve high-quality results.

Furthermore, naively flattening a tri-plane into a 1D sequence of tokens can result in the loss of 3D relative correlations between three planes. 
To preserve the cross-plane correlations in the attention operation, we propose \textit{spatial-aware position embedding} (SAPE). 
Specifically, we create three orthogonal learnable position embeddings and add them to each flattened plane respectively. 
This approach introduces an inductive bias, allowing each token in the flattened tri-plane sequence to differentiate whether other tokens belong to the same plane or two other distinct planes.
In this way, we maintain the 3D relative correlations between planes throughout the model.
More details of linear attention and \textit{spatial-aware position embedding} are introduced in the supplementary material.

\smallskip
\noindent
\textbf{Learning.} 
Our training objective of the spatial-aware auto-encoder can be formulated as follows:
\begin{equation}
    \mathcal{L}_{ae} = \mathcal{L}_{rec} (x, \mathcal{D}(\mathcal{E}(x))) + \mathcal{L}_{KL}(x, \mathcal{D}, \mathcal{E}) + \mathcal{L}_{geo},
\end{equation}
where $\mathcal{L}_{rec}$ is a $L_1$ norm applied between input $x$ and its reconstruction $\mathcal{D}(\mathcal{E}(x))$. To avoid arbitrarily high-variance latent spaces, we introduce a slight KL penalty $\mathcal{L}_{KL}$ towards a standard normal on the learned latent, similar to a VAE~\cite{kingma2013auto}. Furthermore, we additionally add geometry loss $\mathcal{L}_{geo}$ as defined in Eq.~\ref{eq:L_geo} to ensure the faithful representation of the shape in the learned latent tri-plane.

To be noted, the latent tri-plane representation $z$ retains a tri-plane structure with $z = \{z^{(i)} | z^{(i)} \in \mathbb{R}^{c \times n \times n}, i \in \{1, 2, 3\}\}$. Unlike previous work such as DiffusionSDF~\cite{chou2023diffusion} that employs an arbitrary one-dimensional latent vector, our method upholds the inherent 3D structure of the tri-plane representation. By preserving this 3D structure, our compression model is capable of more effectively capturing the details of the input information, contributing to the generation of high-quality 3D shapes.

\subsection{Generative Modeling of Latent Tri-Planes}
\label{sec:ldm}
By employing our innovative spatial-aware tri-plane autoencoder, we can now encode raw tri-planes into a compact, low-dimensional latent tri-plane space. 
This latent representation is more advantageous for likelihood-based generative modeling compared to the raw tri-plane space, as it enables them to focus on the fundamental, semantic aspects of the data in a reduced-dimensional, computationally more manageable space~\cite{rombach2022high}.

\smallskip
\noindent
\textbf{Diffusion Model Formulation.} 
Diffusion Models are probabilistic constructs designed to learn a data distribution $q(z_0)$ by progressively denoising a normally distributed variable.
The learning process is equivalent to performing the reverse operation of a fixed Markov Chain with a length of $T$. The diffusion process transforms latent $z_0$ into purely Gaussian noise
$z_T \sim \mathcal{N} (0,I)$ over $T$ time steps. The forward step in this process is defined as:
\begin{equation}
    q(z_t | z_{t-1})  = \mathcal{N} (z_t; \sqrt{1 - \beta_t} z_{t-1}, \beta_t I),
\end{equation}
where noisy variable $z_t$ is derived by scaling the previous noise sample
$z_{t-1}$ with $\sqrt{1 - \beta_t}$ and adding Gaussian noise following a variance schedule $\beta_1, \beta_2, \dots, \beta_T$. 

\smallskip
\noindent
\textbf{Learning.} 
The goal of training a diffusion model is to learn to reverse the diffusion process. To achieve this, we adopt the approach proposed by Aditya \etal~\cite{ramesh2022hierarchical}, wherein a neural network is used to directly predict $z_0$. Given a uniformly sampled time step $t$ from the set $\{1, ..., T\}$, we generate $z_t$ by sampling noise from the input latent vector $z_0$. A time-conditioned denoising autoencoder~\cite{rombach2022high}, denoted by $\Psi$, is employed to reconstruct $z_0$ from $z_t$. The objective of the latent tri-plane diffusion is given by:
\begin{equation}
    \mathcal{L}_{ldm} = ||\Psi(z_t, \gamma(t)) - z_0||^2,
\end{equation}
where $\gamma(\cdot)$ represents a positional embedding and $|| \cdot ||^2$ denotes the mean squared error (MSE) loss.

During the testing phase, we iteratively denoise $z_T$ to obtain the final output $z'$. This output can be decoded into the raw tri-plane $x'$ with a single pass through the decoder $\mathcal{D}$. Finally, a pretrained MLP decodes $x'$ into a dense signed distance volume using marching cube-shape extraction.

\smallskip
\noindent
\textbf{Condition Injection.} 
To generate novel object with given conditions from various modalities, including point clouds, 2D images, and text descriptions, we inject a latent representation of conditions derived from these modalities. Specifically, for a given input $y$, we utilize a domain-specific encoder $\Upsilon$ to extract shape features  $\pi = \Upsilon(y)$, which subsequently guide the sampling or training process of the diffusion model. 

We employ the denoising U-Net architecture described in Rombach~\etal~\cite{rombach2022high}, augmenting it with an additional cross-attention layer within each block of the U-Net backbone to incorporate condition prompts. The cross-attention layer is defined as:
$
\text{Attention}(Q, K, V) = \text{softmax}\left(\frac{QK^T}{\sqrt{d_k}}\right)V
$, 
with:
\begin{equation}
    Q = W^{(i)}_Q \cdot \psi_i(z_t, \gamma(t)) , \quad K = W^{(i)}_K \cdot \pi, \quad V = W^{(i)}_V \cdot \pi .
\end{equation} 
Here, $ \psi_i(\cdot)$ denotes the output of an intermediate layer of $\Psi$, and $W_Q, W_K, W_V$ represent learnable parameters. 

Furthermore, we adopt the classifier-free guidance paradigm~\cite{ho2022classifier} at each training iteration. That is, we substitute the shape feature with a \textit{zero-mask} as a conditioning input with 10\% probability to enhance sample diversity.

\section{Experiments}
\label{sec:experiment}
Our NeuSDFusion offers a flexible and efficient way to generate high-quality 3D shapes, either unconditionally or guided by different conditions, as demonstrated in the following sections.
We evaluate our approach in various experimental settings, encompassing unconditional content generation, multi-modal shape completion, single-view 3D reconstruction and language-guided generation.

\subsection{Unconditional Generation}

Adhering to the precedent established by previous work~\cite{li2023dqd,vahdat2022lion}, we employ three categories from ShapeNet~\cite{chang2015shapenet}, namely Airplane, Chair and Car, to assess the unconditional generation capability of our proposed method.  
We adopt 1-Nearest Neighbour Accuracy (1-NNA)~\cite{lopez2016revisiting} for evaluating unconditional generation. 1-NNA is a direct measure of distributional similarity, accounting for both diversity and quality.  We measure 1-NNA using both Chamfer Distance (CD)~\cite{cui2023p2c} and Earth Mover Distance (EMD)~\cite{fan2017point} to provide a comprehensive assessment of the  evaluation metric.

\begin{table}
\centering
\caption{Results on \textit{Airplane}, \textit{Chair}, \textit{Car} categories from ShapeNet using 1-NNA$\downarrow$.}
\begin{adjustbox}{width=0.7\textwidth}
\input{tables/unconditional_generation}
\end{adjustbox}
\label{tab:unconditional_generation}
\end{table}

As shown in Tab.~\ref{tab:unconditional_generation}, our method outperforms existing techniques, achieving state-of-the-art results in unconditional generation. Our approach consistently excels across all categories, demonstrating the significant superiority of our proposed pipeline in the detailed modeling of 3D shapes. Notably, we markedly surpass both LION and 3DQD, which also utilize DDPMs but with different object representations. This performance enhancement is largely attributed to our novel object representation method, NeuSDF, and the spatial-aware tri-plane autoencoding scheme. As a result, our samples are diverse and visually appealing, as illustrated in Fig.~\ref{fig:intro}.

\subsection{Multi-Modal Shape Completion}

In accordance with standard practices~\cite{li2023dqd,mittal2022autosdf}, we evaluate the shape completion capability of our proposed method using the ShapeNet dataset, comprising 13 categories and follow the train/test splits provided by DISN~\cite{xu2019disn}. 
The partial shapes encompass two settings: 1) the bottom half of the ground truth, and 2) the octant with front, left, and bottom half of the ground truth. 
For evaluation purposes, we compute the Total Mutual Difference (TMD) of $N = 10$ generated shapes for each input, indicative of completion diversity. 
We report the Minimum Matching Distance (MMD) and Average Matching Distance (AMD), which gauge the minimum and average Chamfer Distances from the ground truth to the $10$ generated shapes, thereby demonstrating completion quality.
\begin{figure}
    \centering
    \includegraphics[width=0.9\linewidth]{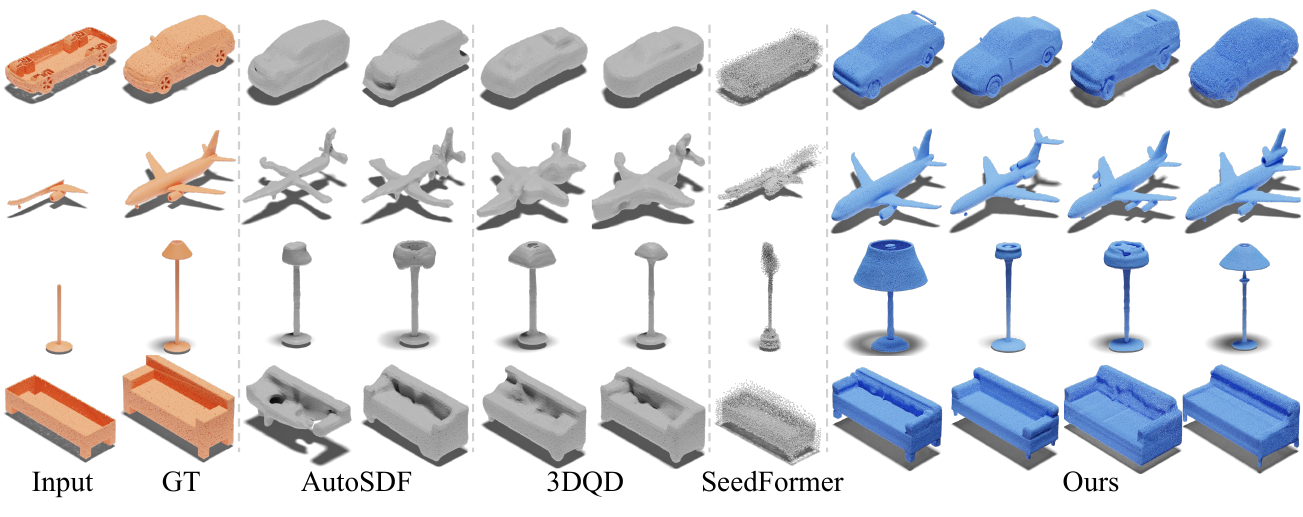}
    \caption{Multi-modal shape completion results. Our NeuSDFusion method generates shapes with superior quality and diversity compared to previous state-of-the-art approaches, while remaining consistent with the input partial shapes.}
    \label{fig:shape-completion}
\end{figure}
\begin{table}
\centering
\caption{Results of multi-modal shape completion, metrics are multiplied by $10^2$.}
\begin{adjustbox}{width=0.6\textwidth}
\input{tables/shape_comp}
\end{adjustbox}
\label{tab:shape_comp}
\end{table}

We compare our approach with both state-of-the-art shape generation~\cite{li2023dqd,mittal2022autosdf} and point cloud completion~\cite{yu2021pointr,zhou2022seedformer} methods. The results are presented in Tab.~\ref{tab:shape_comp}. Notably, our method outperforms the previously best-performing method, 3DQD, in terms of quality metrics for bottom half and octant shape completion, while maintaining a competitive diversity score. Qualitative results in Fig.~\ref{fig:shape-completion} demonstrate that our method exhibits superior performance in diversity while preserving the high-quality geometries. This suggests that our proposed autoencoder effectively embeds raw tri-plane representations into a structural latent space, enabling the subsequent generation process to produce shapes with descriptive conditions.

\subsection{Single-View 3D Reconstruction}
\label{sec:single-view}

\begin{figure}
    \centering
    \includegraphics[width=0.9\linewidth]{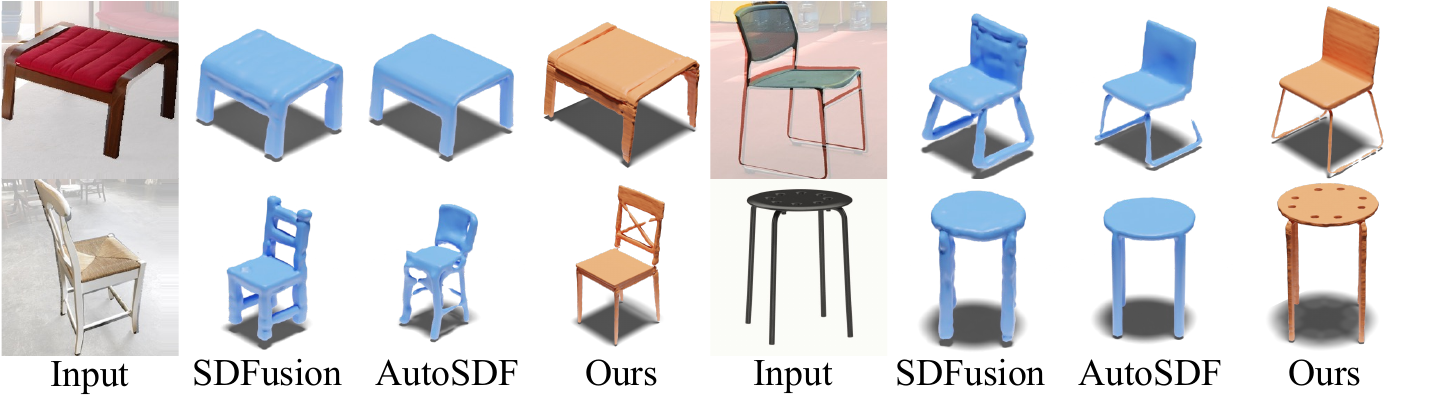}
    \caption{Single-view reconstruction on the Pix3D dataset. Note that our approach generates significantly more detailed shapes compared to previous works, demonstrating the effectiveness of our method in capturing intricate shape properties.}
    \label{fig:single-view}
\end{figure}

Following the precedent set by previous methods~\cite{cheng2023sdfusion, mittal2022autosdf}, we evaluate our approach for 3D shape reconstruction from a single image using the real-world Pix3D~\cite{sun2018pix3d} benchmark dataset. We employ the provided train/test splits for the chair category and, in the absence of official splits for other categories, follow previous methods~\cite{mittal2022autosdf,cheng2023sdfusion} to split samples into train/test sets. We adopt the official evaluation script~\cite{sun2018pix3d} to assess our method and compare it with prior approaches. We contrast our approach with recent state-of-the-art generative single-view reconstruction methods, namely AutoSDF~\cite{mittal2022autosdf} and SDFusion~\cite{cheng2023sdfusion}. The evaluation results are reported in Tab.~\ref{tab:single_view_reconstruction}. Our approach consistently outperforms prior work by a substantial margin, achieving approximately 50\% improvements in both Chamfer Distance and F-Score. Visualizations of our results compared to previous work are presented in Fig.~\ref{fig:single-view}.

Our proposed NeuSDF can effectively represent 3D object shapes, as our pipeline is capable of modeling fine shape details with minimal memory usage, while previous methods are hindered by computational cost. Consequently, the 3D shapes generated by our approach are significantly more detailed than those produced by prior state-of-the-art methods.

\begin{table}
  \centering
  \begin{minipage}{0.35\textwidth}
    \caption{Quantitative results for single-view reconstruction.}
    \centering
    \begin{adjustbox}{width=0.9\textwidth}
    \input{tables/image_generation}
    \end{adjustbox}
    \label{tab:single_view_reconstruction}
  \end{minipage}\hfill
  \begin{minipage}{0.6\textwidth}
    \caption{Results for text-guided generation. PMMD, CLIP-S, and TMD are scaled by $10^2$.}
    \centering
    \begin{adjustbox}{width=0.9\textwidth}
    \input{tables/text_generation}
    \end{adjustbox}
    \label{tab:text_guided}
  \end{minipage}
\end{table}

\subsection{Language-Guided Generation}

To quantitatively evaluate our method on text-guided shape generation, we adopt the approach used in previous work~\cite{li2023dqd, mittal2022autosdf} and utilize the ShapeGlot~\cite{achlioptas2019shapeglot} dataset, which provides text utterances describing the differences between a target shape and two distractors based on the ShapeNet dataset. We employ the same train/test splits provided by AutoSDF~\cite{mittal2022autosdf}. To measure the similarity between text and shape modalities, we follow 3DQD~\cite{li2023dqd} and use CLIP-S as the metric. The metric computes the maximum score of cosine similarity between $N = 9$ generated shapes and their corresponding text prompts using a pre-trained CLIP~\cite{radford2021clip}. Since CLIP cannot directly process 3D shape inputs, we render each generated shape into 20 2D images from various viewpoints to compute CLIP-S. Additionally, we employ Fr\'echet-Pointcloud Distance (FPD)~\cite{shu2019treegcn} and Pairwise Minimum Matching Distance (PMMD) to measure the distance between ground truth and samples.

\begin{figure}[t]
    \centering
    \includegraphics[width=0.9\linewidth]{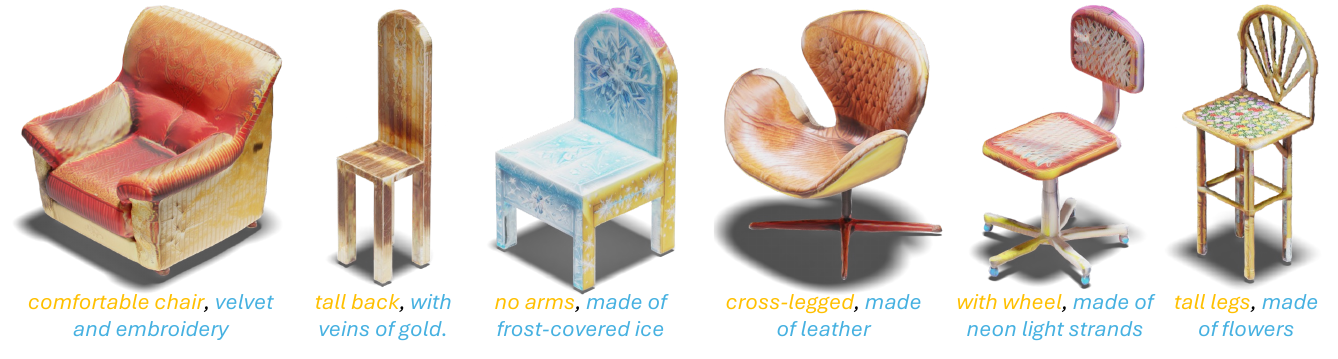}
    \caption{Text-guided shape generation. We showcase the ability of our method \wrt generating shapes based on text prompts. The orange text represents the geometry prompt and we employ Meshy\protect\footnotemark to generate texture using the blue prompt.}
    \label{fig:text}
\end{figure}

Quantitative results in Tab.~\ref{tab:text_guided} demonstrate that our technique consistently outperforms the baseline works across all evaluated metrics, particularly in FPD and PMMD. This indicates that our approach adheres more consistently to text prompts than previous work. We also present text-guided generation results in Fig.~\ref{fig:text}, illustrating that our method can achieve effective modality alignment between text and 3D objects.
\footnotetext{https://www.meshy.ai/}

\subsection{Ablation Study}

\begin{wrapfigure}[10]{r}{0.45\textwidth}
    \vspace{-10pt}
    \centering
    \includegraphics[width=\linewidth]{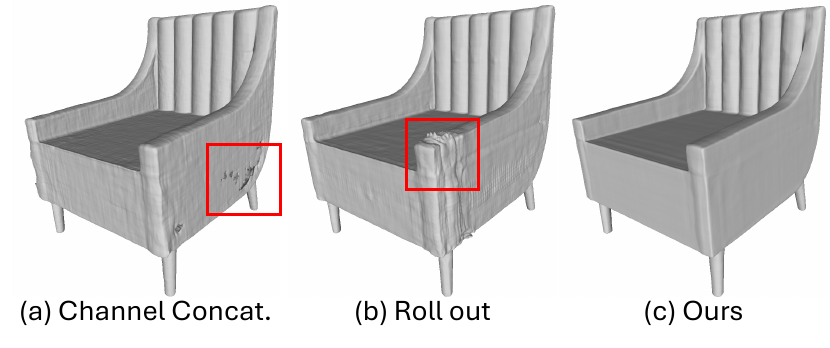}
    \caption{Comparison of different tri-plane autoencoding methods.}
    \label{fig:cnn}
\end{wrapfigure}

We evaluate the effectiveness of our spatial-aware transformer-based autoencoder design by comparing it with two related methods: roll out~\cite{gupta20233dgen,wang2023rodin} and channel-wise concatenation~\cite{shue20233d,chou2023diffusion}. 
Specifically, we train both our autoencoder and a CNN-based autoencoder from Stable Diffusion~\cite{rombach2022high} on a chair-class subset of ShapeNet. As illustrated in Fig.~\ref{fig:cnn}, the reconstructed results of channel-wise concatenation or roll out exhibit defects due to spatial misalignment. 
The roll out method suffers from defects at the border of the tri-plane representation, as the convolution operation overlaps over two planes, while the values at the edge of two adjacent planes are not continuous in space, as shown in Fig.~\ref{fig:cnn} (b). Channel concatenation, in contrast, presents some reconstruction defects due to the lack of an explicit relationship between the same spatial locations across different planes, thereby limiting the reconstruction capability. Our proposed approach employs group convolution to downsample each plane individually and introduces a specially designed attention mechanism to achieve 3D-aware tri-plane interaction. Consequently, our generated objects are integral with a smooth surface. Furthermore, we provide additional evaluation and implementation details of our method in the supplementary material.

\section{Conclusion}
\label{sec:conclusion}

In this paper, we introduce a novel approach to 3D shape synthesis by leveraging a hybrid 3D shape representation and enforcing spatial consistency within the autoencoder module. Our method enables the generation of high-fidelity 3D shapes that are consistent with various condition modalities. Through extensive evaluation, we demonstrate that our proposed approach consistently outperforms previous state-of-the-art methods across multiple evaluation metrics and diverse settings, verifying the effectiveness and robustness of our method. Furthermore, our approach exhibits superior performance in terms of generation diversity and shape quality, highlighting its potential for practical applications in computer vision, graphics, and related fields.

\bibliographystyle{splncs04}
\bibliography{main}
\end{document}

% ---------------------------------------------------------------
\title{NeuSDFusion: A Spatial-Aware Generative Model for 3D Shape Completion, Reconstruction, and Generation \\ 
\large Supplementary Material} 

\titlerunning{NeuSDFusion: A Spatial-Aware Generative Model}

\author{
Ruikai~Cui \inst{1} \thanks{The contribution of Ruikai~Cui, Han~Yan and Zhennan~Wu was made during an internship at Tencent XR Vision Labs.} \and
Weizhe~Liu \inst{2}\and
Weixuan~Sun \inst{2}\and
Senbo~Wang \inst{2}\and
Taizhang~Shang \inst{2}\and
Yang~Li \inst{2} \and
Xibin~Song \inst{2}\and
Han~Yan \inst{3}\and
Zhennan~Wu \inst{4}\and
Shenzhou~Chen \inst{2}\and
Hongdong~Li \inst{1} \and
Pan~Ji \inst{2}
}

\authorrunning{R.~Cui et al.}

% \institute{Australian National University \and
% Tencent XR Vision Labs \and
% Shanghai Jiao Tong University \and
% The University of Tokyo
% }
\institute{
\mbox{%
\inst{1}Australian National University \quad
\inst{2}Tencent XR Vision Labs
}
\mbox{
\inst{3}Shanghai Jiao Tong University \quad
\inst{4}The University of Tokyo
}
}
\maketitle

In this supplementary document, we present additional information and experiments to support the main manuscript. We provide training and architecture details, formulations of evaluation metrics, comprehensive ablation studies, and further visualizations.

\section{Autoencoder Details}
\subsection{Linear Attention}
\begin{figure}
    \centering
    \includegraphics[width=\linewidth]{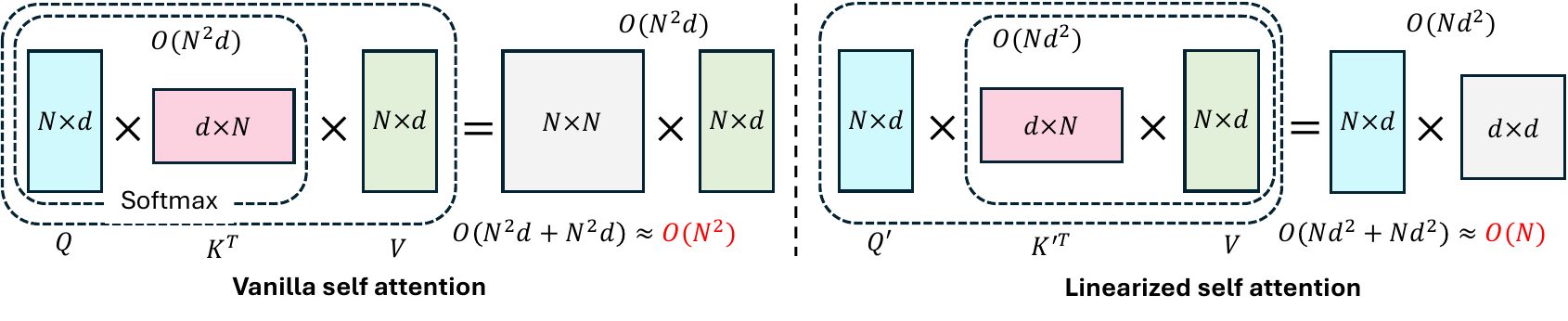}
    \caption{Illustration of the computations for vanilla self attention~\cite{vaswani2017attention} (left) and linearized attention~\cite{qin2022cosformer} (right). The input length is $N$ and feature dimension is $d$, with $d \ll N$. Tensors in the same box are associated for computation. The linear attention formulation allows $O(N)$ time and space complexity.}
    \label{fig:cosformer}
\end{figure}

One of the prominent limitations of the Transformer~\cite{vaswani2017attention} architecture is the computational overhead of the self-attention module. 
This complexity increases quadratically with the sequence length. In our particular setting, the sequence length is $3 \times 64 \times 64 = 12288$ in the first stage of the autoencoder model, which leads to unmanageable computational cost.
This limitation impedes the model's capacity to scale up to higher-resolution tri-planes or larger parameter sizes.
To alleviate the computational cost, we propose leveraging linear attention.
A general attention operation can be formulated as follows:

\begin{equation}
    \mathcal{O}_i
 = \sum_j {\mathcal{S}(Q_i,K_j)}V_j,
    \label{eq: generalized att}
\end{equation}
where $O_i$ denotes the $i$-th token of the output, $Q$, $K$, and $V$ refer to three projected embeddings of the input. 
$\mathcal{S}(\cdot)$ is a function that measures the similarity between queries and values.
In the standard (or 'vanilla') attention mechanism proposed by Vaswani~\etal~\cite{vaswani2017attention}, $\mathcal{S}(\cdot)$ becomes the dot-product attention with softmax normalization, which results in quadratic complexity since an attention matrix has the shape of $N \times N$.
To address the computational overhead associated with the self-attention module, we propose to utilize linear attention. Linear attention eliminates the softmax operation from vanilla attention and employs kernel tricks to achieve linear complexity. Specifically, similarity function in the linear attention can be formulated as follows:

\begin{equation}
\textstyle{
    \mathcal{S}(Q_i, K_j) = \phi(Q_i)\phi(K_j)^T,
    \label{eq: decompose}
}\end{equation}
where $\phi$ is a kernel function used to map the queries and keys to their hidden representations. 
Then, similarities between tokens are obtained via simple matrix multiplication. 
The choice of kernel function can vary depending on the specific implementation. For instance, 
$\phi$ is $elu+1$ in ~\cite{katharopoulos2020transformers}, $cosine$ in \cite{qin2022cosformer}, Gaussian approximation in \cite{choromanski2020rethinking} and $Relu$ in ~\cite{qin2022devil}. 
This decomposition allows achieving linear complexity through the matrix multiplication property, as follows:
\begin{equation}
\textstyle{
    (\phi (Q) \phi({K})^T) V = \phi (Q) (\phi({K})^T V).
    \label{eq: linear form}
}\end{equation}
In this form (Eq.~\ref{eq: linear form}), we can avoid computing the attention matrix $A = QK^T \in\mathbb{R}^{N \times N}$ by first calculating the $\phi({K})^T V \in\mathbb{R}^{d \times d}$ and then multiplying it by $\phi(Q)\in\mathbb{R}^{N \times d}$. 
This approach reduces the computation complexity to $O(Nd^2)$. 
Fig.~\ref{fig:cosformer} provides a conceptual illustration of linear attention. 
As shown, the matrix multiplication order is changed in the attention calculation and the attention matrix $A = QK^T \in\mathbb{R}^{N \times N}$ is avoided.

In our practical implementation, we adopt the the linear attention proposed in ~\cite{qin2022devil,qin2023scaling}.

\subsection{Spatial Aware Positional Embedding}
The Transformer architecture is designed to perform spatial understanding (token mixing operation) on a 1D sequence in the attention module. Therefore, to apply this structure to tri-plane data, we need to flatten the tri-plane into a 1D sequence of tokens $x \in \mathbb{R}^{C \times 3HW}$.
However, this approach has a drawback in that it loses plane-wise 3D information, as each token is not able to determine whether other tokens are from the same plane or two other planes.
To preserve cross-plane correlations in the attention operation, 
a specific position embedding could be incorporated.
The position embedding should have two properties: 1) it should help the attention module to preserve plane-wise information, and 2) it should be compatible with linear attention.

To this end, we propose a novel approach called \textit{spatial-aware position embedding} (SAPE). 
We begin by considering the flattened 1D sequence of the tri-plane, denoted as $x = [0:HW-1, HW:2HW-1, 2HW:3HW-1] \in \mathbb{R}^{C \times 3HW} $, which is the concatenation of three flattened planes. 
We then initialize 3 position embeddings $p \in \mathbb{R}^{C \times 3 \times HW} $ and add them to each plane, respectively.
Furthermore, since the three planes are orthogonal to each other in the tri-plane, we aim to introduce a similar inductive bias into the attention calculation.
Therefore, we force the three position embeddings to be orthogonal to each other so the product of embeddings from different planes will be 0.
Specifically, we leverage Householder transformation to achieve mutual orthogonalization of three embeddings.
We create two learnable vectors $h \in \mathbb{R}^{C} $ and $t \in \mathbb{R}^{HW}$, three position embeddings are obtained via:
\begin{equation}
\begin{aligned}
    H & = I - \frac{2}{h^{T}h}hh^{T} \in \mathbb{R}^{C \times C} \\
    h_i & = H[:,i] \in \mathbb{R}^{C \times 1} , (i=0,1,2)  \\
    p_i & = h_i \times t \in \mathbb{R}^{C \times HW}, (i=0,1,2)  
\end{aligned}
\end{equation}
where $H\in \mathbb{R}^{C \times C}$ is a Householder matrix generated from the learnable vector $h$.
and its each column is mutually orthogonal~\cite{petersen2008matrix}. 
We extract the first three columns and transform them into three position embeddings \ie, $p_1, p_2, p_3$, by multiplying a learnable vector $t$. 
To this end, we add the three position embeddings to each flattened plane in an element-wise manner to achieve spatial-aware self-attention.
We ablate SAPE in Sec.~\ref{section ablation sape}.

\section{Background on Diffusion Probabilistic Models}

Diffusion Models are probabilistic models designed to learn a data distribution $z_0 \sim q(z_0)$ by gradually denoising a normally distributed variable. This process corresponds to learning the reverse operation of a fixed Markov Chain with a length of $T$. The inference process works by sampling a random noise $z_T$ and gradually denoising it until it reaches a meaningful latent $z_0$. Denoising diffusion probabilistic model (DDPM)~\cite{Ho2020ddpm} defines a diffusion process that transforms latent $z_0$ to white Gaussian noise $z_T \sim \mathcal{N} (0,I)$ in $T$ time steps. Each step in the forward direction is given by:
\begin{align}
   & q(z_1, ..., z_T | z_0)  = \prod_{t=1}^{T} q(z_t | z_{t-1}) \\
   & q(z_t | z_{t-1}) = \mathcal{N} (z_t; \sqrt{1 - \beta_t}z_{t-1}, \beta_t I)
\end{align}
The noisy latent $z_t$ is obtained by scaling the previous noise sample $z_{t-1}$ with $\sqrt{1 - \beta_t}$ and adding Gaussian noise with variance $\beta_t$ at timestep $t$. During training, DDPM reverses the diffusion process, which is modeled by a neural network $\Psi$ that predicts the parameters $\mu_{\Psi}(z_t, t)$ and $\Sigma_{\Psi}(z_t, t)$ of a Gaussian distribution.
\begin{equation}
p_{\Psi}(z_{t-1} | z_t) = \mathcal{N} (z_{t-1}; \mu_{\Psi}(z_t, t), \Sigma_{\Psi}(z_t, t))
\end{equation}
With $\alpha_t := 1 - \beta_t$ and $\bar{\alpha} := \prod_{s=0}^{t} \alpha_s$, we can write the marginal distribution:
\begin{align}
& q(z_t | z_0) = \mathcal{N} (z_t; \bar{\alpha}_t z_0, (1 - \bar{\alpha}_t)I) \\
& z_t = \bar{\alpha}_t z_0 + \sqrt{1 - \bar{\alpha}_t}\epsilon
\end{align}
where $\epsilon \sim \mathcal{N} (0,I)$. Using Bayes' theorem, we can calculate $q(z_{t-1} | z_t, z_0)$ in terms of $\tilde{\beta}_t$ and $\tilde{\mu}_t (z_t, z_0)$ which are defined as follows:
\begin{align}
& \tilde{\beta}_t := \frac{1 - \bar{\alpha}_{t-1}}{1 - \bar{\alpha}_t} \beta_t \\
\label{eq:mu}
& \tilde{\mu}_t (z_t, z_0) := \frac{\sqrt{\bar{{\alpha}}_{t-1}}\beta_t}{1 - \bar{\alpha}_t} z_0 + \frac{\sqrt{\alpha_t} (1 - \bar{\alpha}_{t-1})}{1 - \bar{\alpha}_t} z_t \\
& q(z_{t-1} | z_t, z_0) = \mathcal{N} (z_{t-1}; \tilde{\mu}_t (z_t, z_0), \tilde{\beta}_t I)
\end{align}
Instead of predicting the added noise as in the original DDPM, in this paper, we predict $z_0$ directly with a neural network $\Psi$ following Aditya \etal~\cite{ramesh2022hierarchical}. The prediction could be used in Eq.~\ref{eq:mu} to produce $\mu_{\Psi}(z_t, t)$. Specifically, with a uniformly sampled time step $t$ from $\{1, ...,T\}$, we sample noise to obtain $z_t$ from input latent vector $z_0$. A time-conditioned denoising auto-encoder $\Psi$ learns to reconstruct $z_0$ from $z_t$. The objective of latent tri-plane diffusion reads
\begin{equation}
\mathcal{L}_{ldm} = \|\Psi(z_t,\gamma(t)) - z_0\|^2
\end{equation}
where $\gamma(\cdot)$ is a positional encoding function and $\|\cdot\|^2$ denotes the Mean Squared Error loss.

\section{Implementation Details}
\label{sec:implementation}

\subsection{NeuSDF Fitting}

Our NeuSDFusion approach handles each object through a NeuSDF representation, consisting of a tri-plane that stores geometric features and a Multi-Layer Perceptron (MLP) that decodes these features into a real value. For dataset processing, each 3D object is associated with a unique tri-plane, while feature decoding is performed by a dataset-wide shared MLP.
The process begins with a joint training phase, during which we initially train a shared SDF MLP in conjunction with 50 objects from the dataset. 
Upon convergence, the MLP is frozen and considered as a generalizable SDF decoder. 
Subsequently, Following this, we optimize the tri-planes for the remaining objects in the dataset. 
 All tri-planes beyond the initial 50 are individually optimized using the same shared MLP, allowing for efficient parallelization. Each object requires 8 minutes on an NVIDIA Tesla T10 for this process.

We adopt the MLP initialization trick introduced by SAL~\cite{atzmon2020sal}, which constrains the initial SDF output to approximate a rough sphere. 
This spherical geometry initialization technique significantly facilitates global convergence. Empirically, we set the loss weights to $\lambda_1 = 100.0$, $\lambda_2 = 3.0$, $\lambda_3 = 1.0$, and $\lambda_4 = 0.5$ across all datasets. 
In this work, the output tri-plane has a resolution of $H = W = 128$ and $C = 32$ channels.
 
\subsection{Condition Encoders}
To incorporate custom conditions into the generation process, we convert conditions from various modalities into a latent representation. 
We experiment with three modalities: partial point clouds, 2D images, and text.
We use the CLIP~\cite{radford2021clip} encoder for text and images, embedding conditions into a 1024-dimensional latent space.
For the point cloud completion task~\cite{yuan2018pcn,cui2022energy}, we encode partial point clouds with a PoinTr~\cite{yu2021pointr} fine-tuned on benchmark datasets and use the global representation from the PoinTr encoder as the prompt embedding. 
This encoder also processes each partial object into a latent representation of $\mathbb{R}^{1024}$.

\subsection{Network Training and Sampling}
Both the autoencoder and diffusion model are trained on NVIDIA V100 GPUs, requiring 96 and 5 V100 GPU hours, respectively, for a dataset of 5k objects. In terms of the autoencoder implementation, we employ a tri-plane of size $3\times32\times128\times 128$. The encoder down-sampling rate is $f = H/h = W/w = 128/8=16$, and the latent feature dimension is 4, yielding a compressed latent representation of size $3\times4\times8\times8$.

To generate samples from our model, we employ the DDIM~\cite{song2020denoising} sampler with 50 denoising steps. We then use the decoder to transform the sampled latent representation into a tri-plane. Furthermore, we apply the Marching Cubes~\cite{lorensen1998marching} technique to extract 3D objects from the decoded tri-planes. To accomplish this, we further interpolate the tri-planes of spatial resolution  $128 \times 128$ as a $512^3$ distance field for marching cubes.

\section{Evaluation Metrics}

We provide a detailed formulation of metrics that are not explicitly defined in the main manuscript to enhance understanding of our experimental outcomes.

For unconditional generation, we adopt the 1-nearest neighbor accuracy (1-NNA) as the metric, in line with previous works~\cite{vahdat2022lion,li2023dqd}. Let $S_g$ be the set of generated meshes and $S_r$ the set of reference meshes. For evaluation, we calculate the distance between point clouds, from which we sample 2,048 points on each mesh's surface. We generate the same number of samples as the reference set, \ie, $|S_g| = |S_r|$. 
1-NNA measures the similarity between two shape distributions, offering an intuitive assessment where a 1-NN classifier attempts to classify each sample as originating from either $S_r$ or $S_g$. Ideally, if the generated and reference sets are indistinguishable, classifier accuracy approaches 50\%.
Let $S_c = S_r \cup S_g \setminus \{c\}$ and $N_c$ the nearest neighbor of $c$ in $S_c$.
\begin{equation}
    \text{1-NNA} (S_g, S_r) = \frac{\sum_{c \in S_g} I[N_c \in S_g] + \sum_{Y \in S_r} I[N_Y \in S_r]}{|S_g| + |S_r|},
\end{equation}
where $I$ is the indicator function.

For multi-modal shape completion evaluation, we follow 3DQD~\cite{li2023dqd} and adopt Minimum Matching Distance (MMD), 
Average Matching Distance (AMD) and Total Mutual Difference (TMD) as evaluation metrics. From each reference point cloud $P_i \in S_r$, we obtain a partial point cloud $p_i \subset P_i$. From $p_i$, we generate $k = 10$ samples, and denote $c_{ij}$ as the $j$-th predicted mesh to complete $p_i$. Thus, $S_g = \{\sum_{i=1}^{|S_r|} \sum_{j=1}^k c_{ij}\}$ and $|S_g| = k|S_r|$. We use Chamfer Distance (CD)~\cite{qi2017pointnet,wu2021density,cui2023p2c} as the distance measure. We provide definition of the above metrics as below: 

\noindent\textbf{Minimum Matching Distance (MMD)}. For each point cloud in $S_r$, we compute its distance to its nearest neighbor in $S_g$. MMD measures quality with a score that is low when the generated set consists of point clouds close to those in the reference set.
\begin{equation}
\text{MMD} (S_g, S_r) = \frac{1}{|S_r|} \sum_{Y \in S_r} \min_{c \in S_g} \text{CD}(c, Y),
\end{equation}
where CD is the Chamfer Distance.

\noindent\textbf{Average Matching Distance (AMD)} measures the averaged distance of samples in $S_r$ to the set $S_g$. Specifically, for each point cloud in $S_r$, we compute the average distance over its $k=10$ predictions in $S_g$.
\begin{equation}
\text{AMD} (S_g, S_r) = \frac{1}{k|S_r|} \sum_{i \in |S_r|} \sum_{j=1}^{j \leq k} \text{CD}(c_{ij}, P_i).
\end{equation}

\noindent\textbf{Total Mutual Difference (TMD)} is defined as the average pairwise distance among all generated samples for a given conditional input, that is
\begin{equation}
\text{TMD}(S_g) = \frac{1}{|S_g|} \sum_{i=1}^{|S_g|} \left[ \sum_{j=1}^k \frac{1}{k-1}  \sum_{\substack{1\leq l\leq k \\ l \neq j}}^k \text{CD} (c_{ij}, c_{il}) \right].
\end{equation}

Additionally, we adopt \textbf{Coverage (COV)} as a metric in our ablation study. COV measures the proportion of unique nearest neighbors in $S_g$ for each point cloud in $S_r$, serving as an indicator of mode collapse. If mode collapse occurs, the reference point clouds will map to a small set of nearest neighbors, resulting in a low COV score. COV is defined as:

\begin{equation}
    \text{COV} (S_g, S_r) = \frac{|\{\argmin_{Y \in S_r} \text{CD}(c, Y) | c \in S_g\}|}{|S_r|}.
\end{equation}

\section{Additional Ablation Studies}

% \subsection{Effectiveness of \textit{All as Token} strategy}

% We evaluate the  design by comparing it with two related methods: roll out~\cite{gupta20233dgen,wang2023rodin} and channel-wise concatenation~\cite{shue20233d,chou2023diffusion}. Specifically, we train our autoencoder and a CNN-based autoencoder from Stable Diffusion~\cite{rombach2022high} using these methods on a chair-class subset of ShapeNet. As illustrated in Fig.~\ref{fig:cnn}, the reconstructed results of channel-wise concatenation or roll out exhibit defects due to spatial misalignment. The roll out method suffers from defects at the border of the tri-plane representation, as the convolution operation overlaps two planes,
% \begin{wrapfigure}[10]{r}{0.45\textwidth}
%     \centering
%     \includegraphics[width=\linewidth]{figures/cnn.pdf}
%     \caption{Comparison of different tri-plane autoencoding methods.}
%     \label{fig:cnn}
% \end{wrapfigure}
% while the values at the edge of two adjacent planes are not continuous in space, as shown in Fig.~\ref{fig:cnn} (b). Channel concatenation, in contrast, presents some reconstruction defects due to the lack of an explicit relationship between the same spatial locations across different planes, thereby limiting the reconstruction capability. Our proposed approach employs group convolution to downsample each plane individually and introduces a specially designed attention mechanism to achieve 3D-aware tri-plane interaction. Consequently, our generated objects are integral with a smooth surface. 

\subsection{Effectiveness of Spatial-Aware Autoencoder}
\begin{table}
\centering
\caption{Comparison results on shape reconstruction metrics of different tri-plane processing choices. CD and EMD are scaled by $10^3$ and $10^2$, respectively.}
\input{supplementary/tables/planes}
\label{tab:planes}
\end{table}

In contrast to previous tri-planes processing methods, such as roll out or channel concatenation, we designed a novel transformer-based autoencoder that treats all plane entries as tokens, thereby maintaining both in-plane and cross-plane communication. 
We adhere to the same setting as the main manuscript to verify the performance of our proposed autoencoder, utilizing a subset of 32 objects from the ShapeNet chair category for evaluation. 
We then train three different versions of the tri-plane autoencoders on the same chair samples and provide the quantitative results in Tab.~\ref{tab:planes}, using Chamfer Distance (CD) and Earth Mover Distance (EMD) as evaluation metrics. 
Our observations indicate that by treating all plane entries as tokens, and processing tri-planes through our transformer-based autoencoder, we achieve significant improvements over existing methods.

\subsection{Effectiveness of Spatial-Aware Positional Embedding}
\label{section ablation sape}
\begin{table}
\centering
\caption{Ablation study of SAPE. CD and EMD are scaled by $10^3$ and $10^2$, respectively.}
\input{supplementary/tables/sape}
\label{tab:sape}
\end{table}
\begin{figure}
    \centering
    \includegraphics[width=\linewidth]{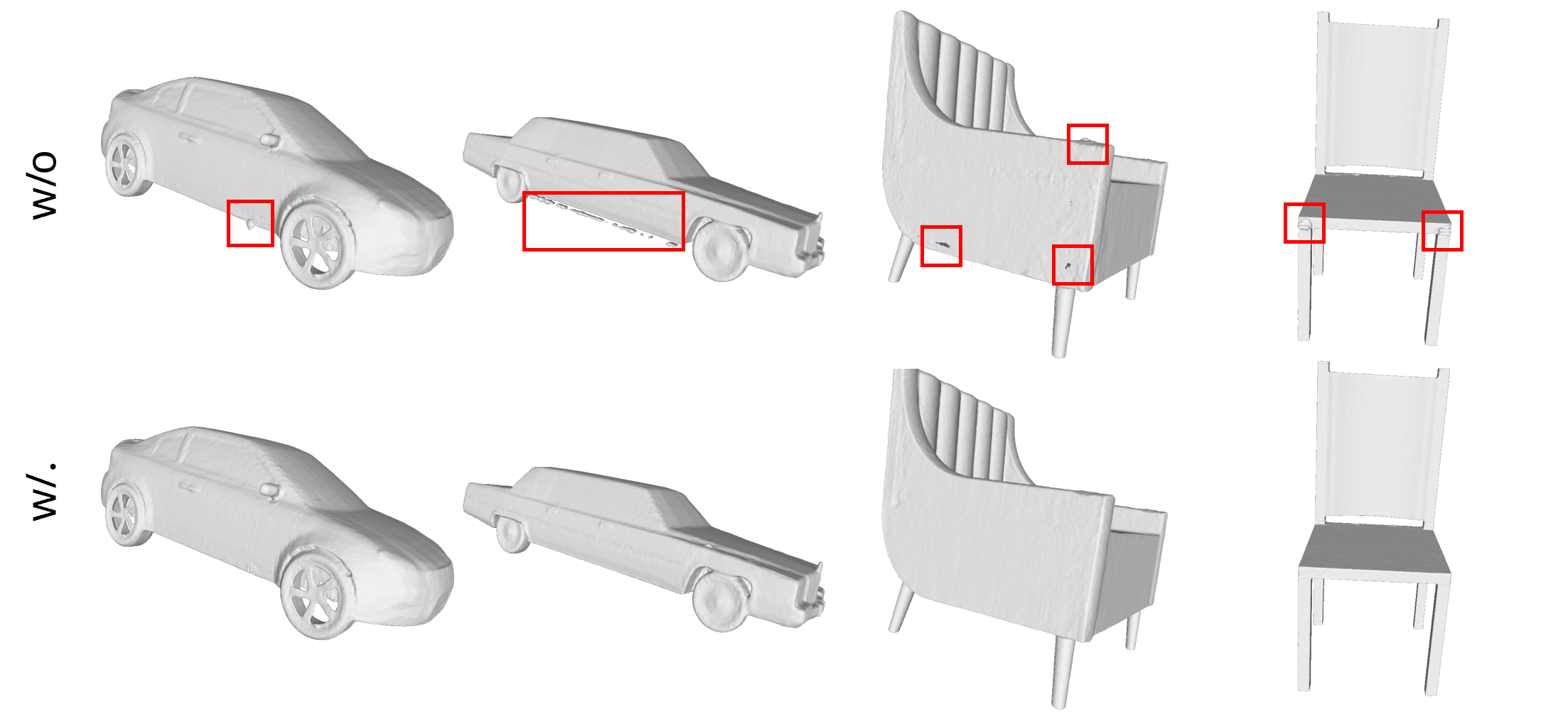}
    \caption{Qualitative comparison between models with and without SAPE, with artifacts highlighted in red boxes.}
    \label{fig:sape}
\end{figure}

We conduct an ablation study to validate the effectiveness of the proposed Spatial-Aware Positional Embedding (SAPE). 
To achieve this, we remove the SAPE module from our designed autoencoder and rely solely on the self-attention mechanism within the transformer blocks to learn the relationships between tri-plane tokens. 
we train both models on 8 tri-planes for the same number of iterations to ensure a focused comparison.
Subsequently, we assess the quality of the reconstructed tri-planes by extracting 3D objects from those planes and measuring the distance between the ground truth objects and the extracted objects. 
The quantitative results are presented in Tab.~\ref{tab:sape}, which demonstrates that SAPE leads to better reconstruction quality. Fig.~\ref{fig:sape} illustrates the benefit of our proposed SAPE module 
% \RK{to be drawn}, 
from which we can observe that reconstruction results with SAPE have more details and fewer artifacts (floater, incomplete surface) than the counterpart. 
This improvement occurs as SAPE explicitly provides inter-plane and intra-plane spatial relationships, allowing the autoencoder to capture shape details more accurately.

\subsection{Effectiveness of Latent Diffusion}

\begin{table}
\centering
\caption{Comparison results between raw tri-plane diffusion and latent space diffusion.}
% \begin{adjustbox}{width=0.4\textwidth}
\input{supplementary/tables/ldm}
% \end{adjustbox}
\label{tab:ldm}
\end{table}

\begin{figure}
    \centering
    \includegraphics[width=\linewidth]{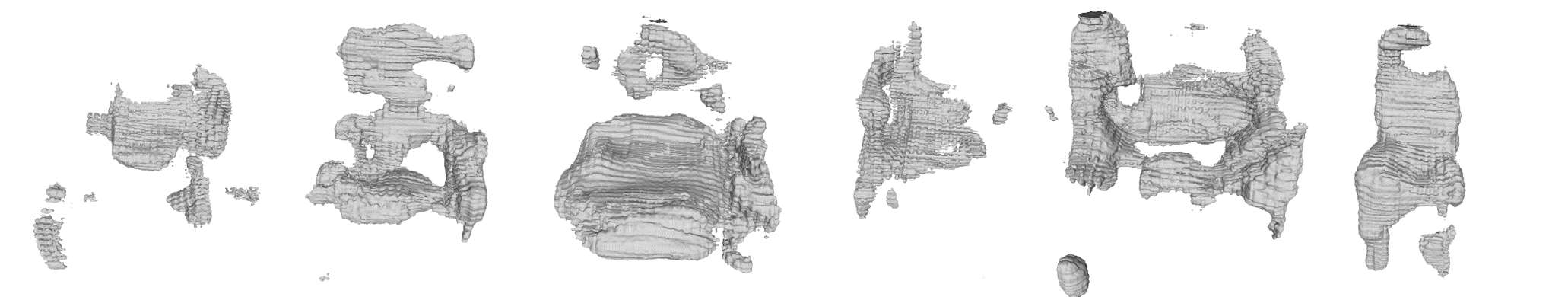}
    \caption{Qualitative results of unconditioned shape generation using raw tri-plane diffusion.}
    \label{fig:raw_diff}
\end{figure}

We emphasize the importance of compressing an object's raw tri-planes into a latent representation in the second stage. To this end, we use the denoising U-Net described in the main paper, testing one model on raw data space and another on latent space.
Both models are trained on the entire ShapeNet Chair category until reaching the same loss convergence threshold.
We find that, unlike the image data structure where raw data space diffusion can still function with some efficiency and quality reduction, denoising diffusion models struggle to adequately model the distribution of raw tri-planes. 
This inefficiency stems from raw tri-planes being stored with an excessive number of parameters and the unconstrained variance in the tri-plane parameter space. Consequently, a raw space diffusion model faces difficulties in generating valid 3D shape NeuSDF representations.
The quantitative results presented in Tab.~\ref{tab:ldm} illustrate superior evaluation metrics for latent space diffusion over raw tri-plane diffusion.
Fig.~\ref{fig:raw_diff} depicts the shapes generated from raw tri-plane diffusion, which exhibit inferior quality compared to our latent space diffusion results in the main paper. Both quantitative and qualitative results demonstrate the necessity of compressing raw tri-planes to a latent representation.

\subsection{Effectiveness of NeuSDF Representation}
\begin{table}
\centering
\caption{Comparison results on shape metrics of different 3D representation settings.}
\input{supplementary/tables/neusdf}
\label{tab:neusdf}
\end{table}

\begin{figure}[t]
    \centering
    \includegraphics[width=\linewidth]{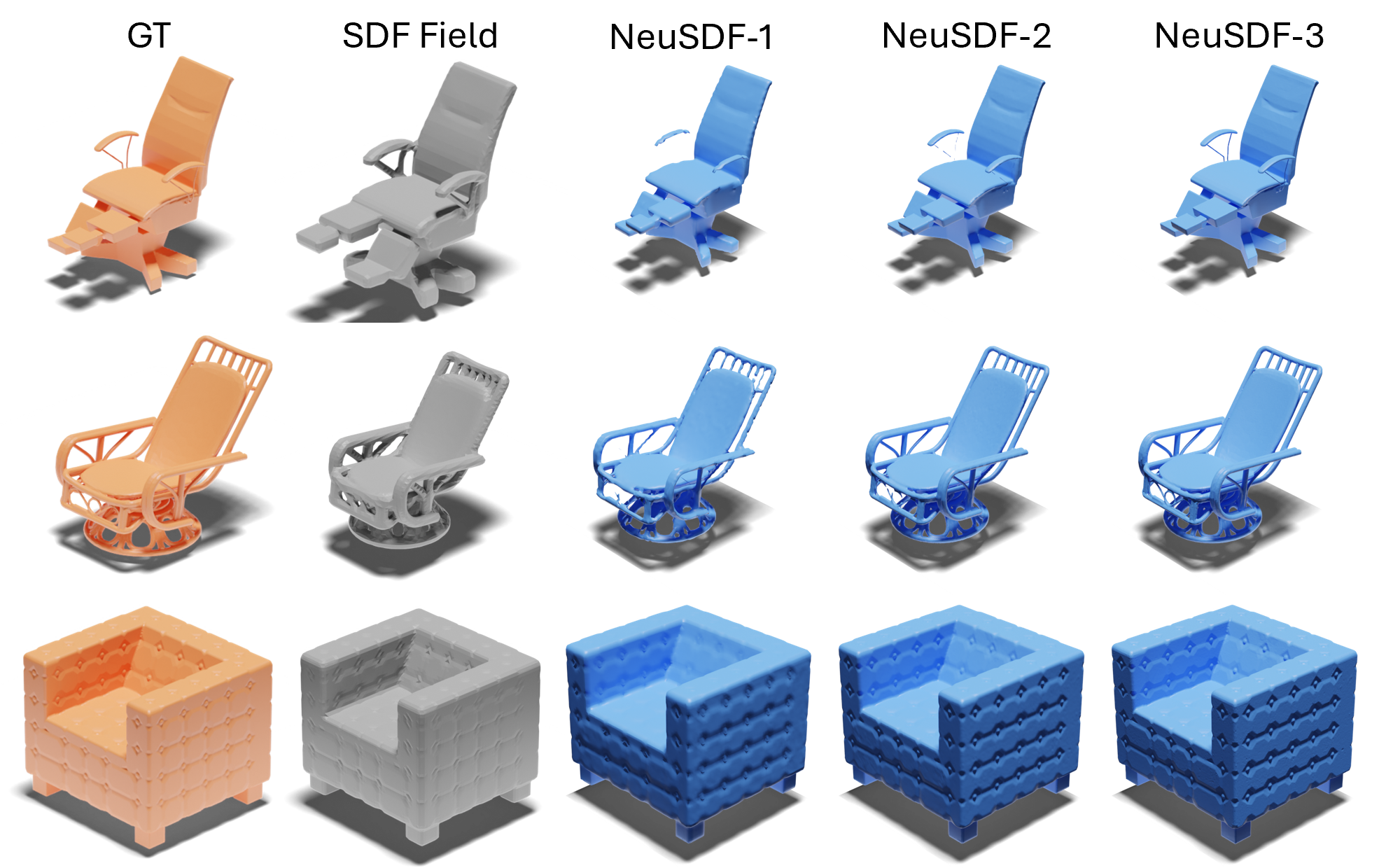}
    \caption{Qualitative results on different shape representation.}
    \label{fig:neusdf}
\end{figure}

This section evaluates the effectiveness of our novel NeuSDF representation in enhancing generation quality.
Our NeuSDF representation decomposes a 3D object into a tri-plane that stores geometric features and an MLP that decodes the feature into a signed distance value. 
This is distinct from previous SDF representations that implicitly embed an object as a neural network~\cite{park2019deepsdf} or explicitly store signed distances as a distance field~\cite{mittal2022autosdf,li2023dqd}. As EG3D~\cite{chan2022efficient} discusses, the implicit neural network-based SDF representation queries slowly, requiring a full MLP pass for each position. Furthermore, distance field variants scale poorly with resolution, limiting their practicality. Previous generative methods~\cite{li2023dqd, park2019deepsdf, cheng2023sdfusion} typically choose distance field-based SDF as the output from the generative model since it can be easily processed by 3D convolutional neural networks. However, the scalability of this explicit SDF representation is limited by its resolution. For instance, SDFusion and 3DQD both use a voxel resolution of $64^3$ due to the memory consumption of higher resolution representations. 
As a result, although the surface of objects retrieved from this SDF representation is still smooth, which is an advantage of SDF representation, finer details and thin structures are absent.

Our qualitative results in Fig.~\ref{fig:neusdf} validate these points. We observe that thin structures extracted from the isosurface of distance field-based SDF are replaced by much coarser geometries, while our method maintains such structures even at the same interpolation resolution.
Notably, in the NeuSDF-3 variant of our proposed representation, we store the tri-planes as a $128\times 128\times 3$ spatial resolution and interpolate it as a $512 \times 512 \times 512$ distance field. Our result preserves most details of the ground truth shape. We also provide a memory consumption and reconstruction error comparison in Tab.~\ref{tab:neusdf}. In comparison, when constrained to a $64\times 64$ resolution, our NeuSDF occupies only 25\% of the space required by voxel SDF representations while significantly reducing reconstruction error. When we extract an object from a $128 \times 128$ plane and interpolate the plane to $512\times 512$, we can achieve the best reconstruction quality indicated by the error metric. It should be noted that part of the reason the SDF field representation results in a much higher error than ours is the loss of fine details, and thin structures are extracted as coarse parts, as shown in Fig.~\ref{fig:neusdf}.

\subsection{Shape Novelty Analysis}

\begin{figure}
    \centering
    \includegraphics[width=0.75\linewidth]{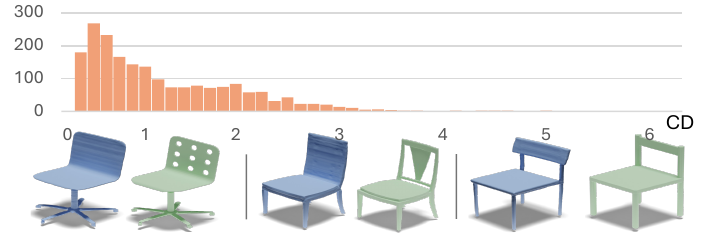}
    \caption{Top: Histogram on the distribution of CD ($\times 100$) between generated chairs and training shapes. Bottom: The nearest shapes (green) retrieved from the training dataset according to their Chamfer distance to the generated chair (blue).}
    \label{fig:novel}
\end{figure}

To statistically analyze the novelty of the generated shapes, we visualize the distribution of Chamfer Distance between our generated shapes and the nearest neighbour in the training shapes in Fig.~\ref{fig:novel}, demonstrating that our method generates novel objects rather than merely memorizing the training data.

\section{Limitations and Future Works}
Despite the high-quality 3D generation results we have demonstrated, our method still has a few limitations. Firstly, our NeuSDFusion utilizes the signed distance function as the underlying representation of 3D objects, which only applies to watertight objects. As a result, our method cannot be used for generating open surfaces, such as clothes. Secondly, our method only generates the geometry of 3D objects without any textural information, while textures are necessary for realistic 3D generation.

\section{Additional Qualitative Results}

\begin{figure}
    \centering
    \includegraphics[width=\linewidth]{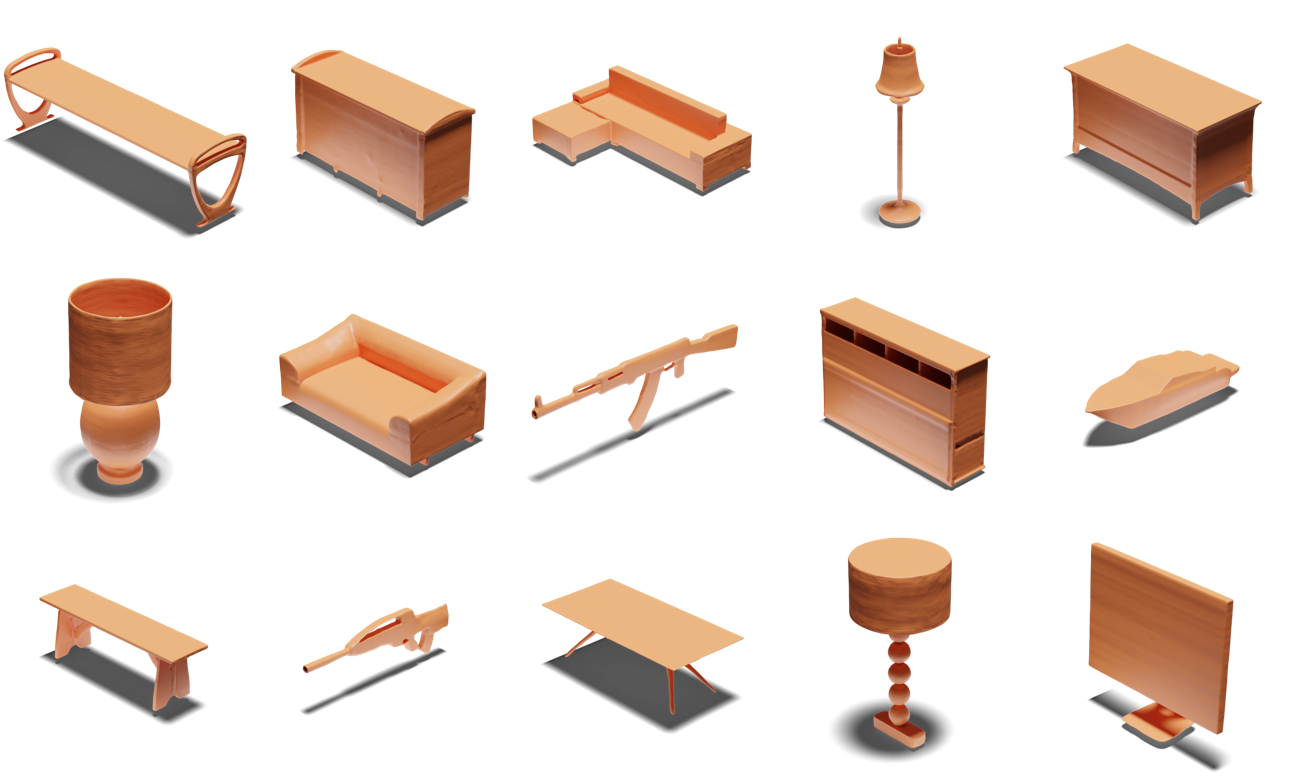}
    \caption{Unconditional generation samples from our proposed method trained on 13 ShapeNet classes. Our method produces clean meshes with detailed geometries and diverse structures.} 
    \label{fig:sup_multiclass}
\end{figure}

\begin{figure}
    \centering
    \includegraphics[width=\linewidth]{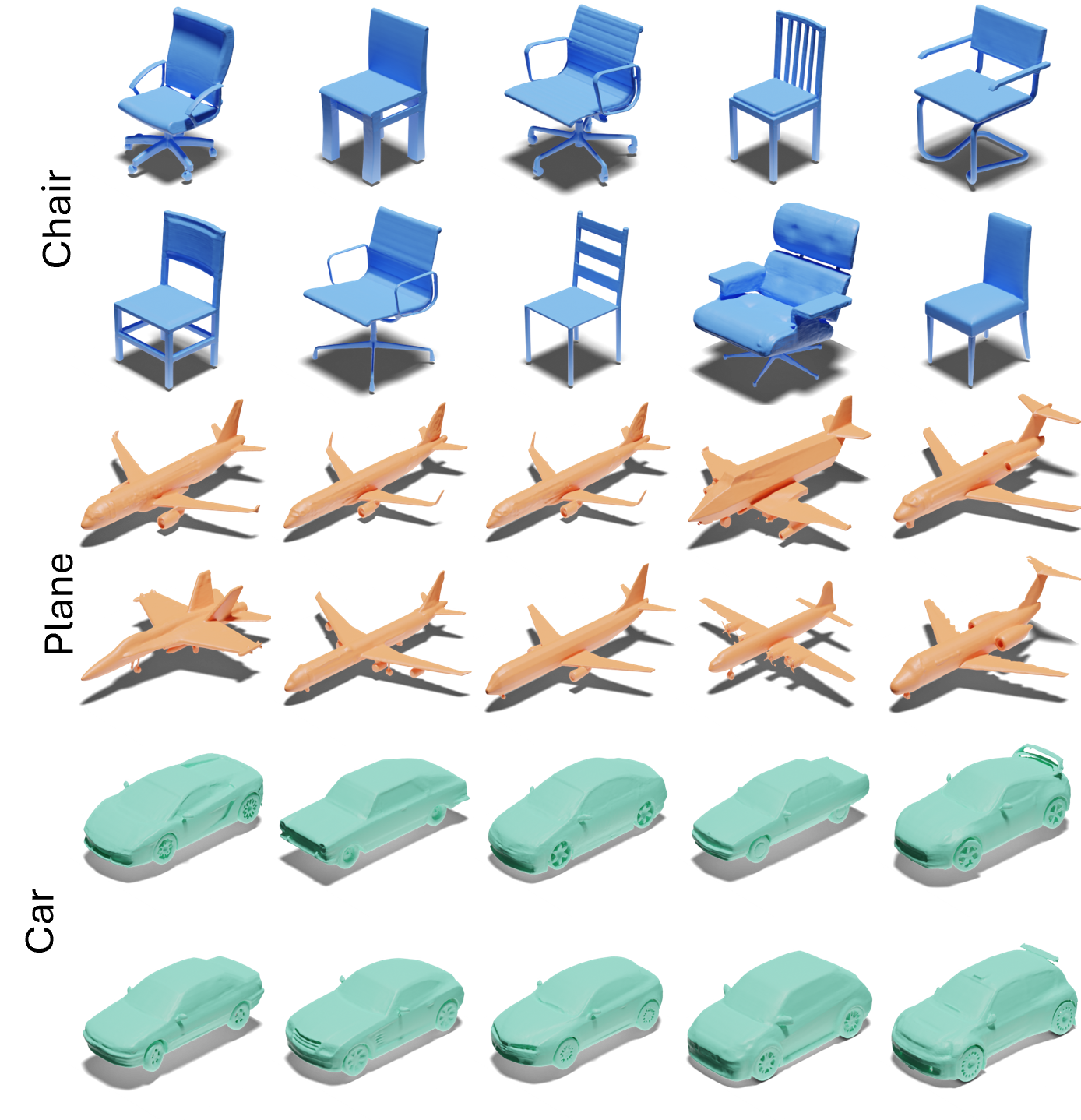}
    \caption{Unconditional generation samples from our proposed method trained on Chair, Plane, Car categories, respectively. Our method produces clean meshes with detailed geometries and diverse structures.} 
    \label{fig:sup_single}
\end{figure}

\begin{figure}
    \centering
    \includegraphics[width=0.95\linewidth]{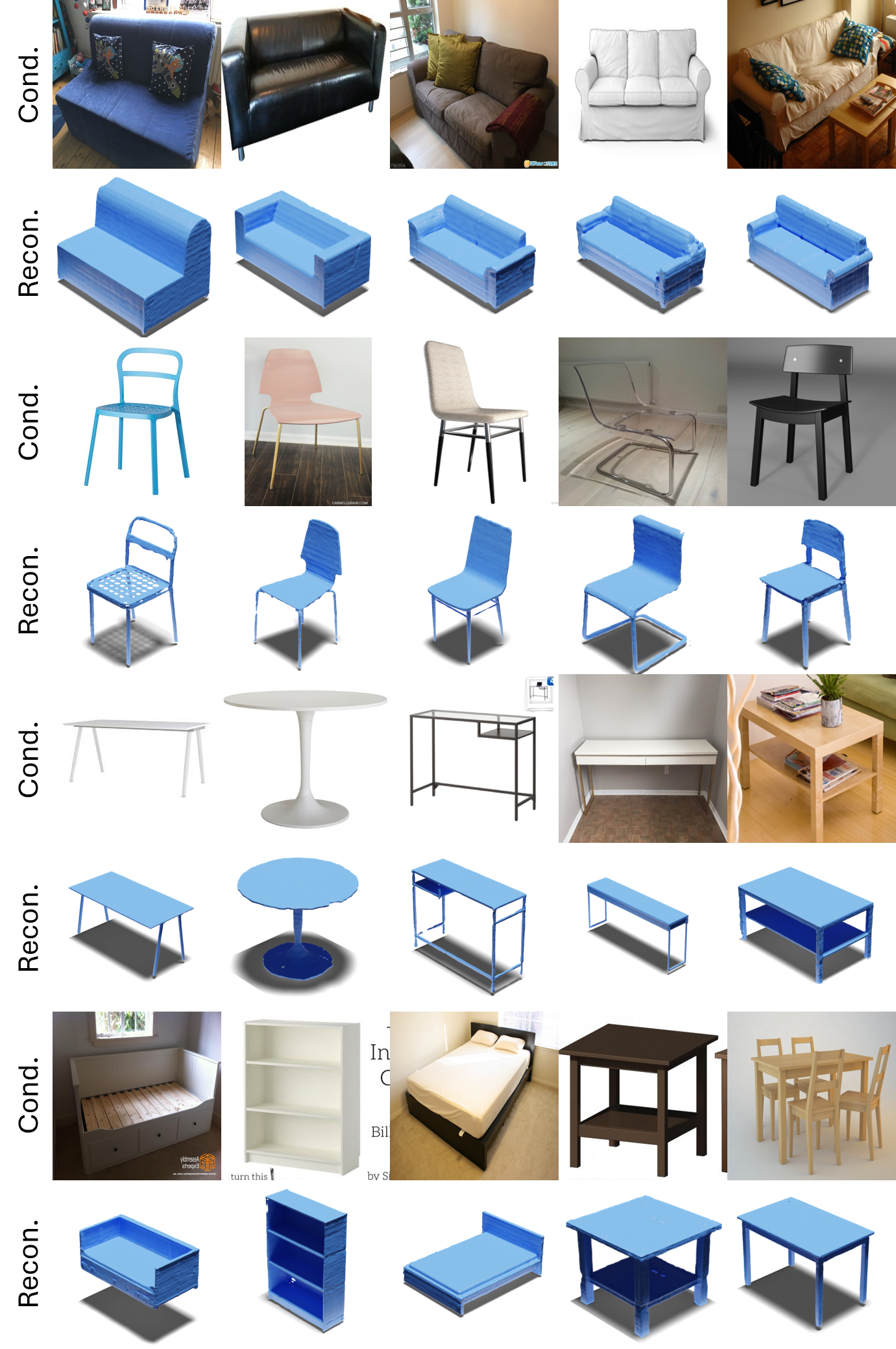}
    \caption{Additional generations guided by images for single-view reconstruction.}
    \label{fig:sup_pix3d}
\end{figure}

To further illustrate the capabilities of our model, we include additional results for unconditional generation in Fig.~\ref{fig:sup_single} and \ref{fig:sup_multiclass}. Additionally, we present guided generation results in Fig.~\ref{fig:sup_pix3d}, showcasing the versatility of our approach.

\clearpage
\bibliographystyle{splncs04}
\bibliography{supplementary/reference}

%% file: tables/unconditional_generation.tex
\begin{tabular}{lccccccc}
\toprule
       & & \multicolumn{2}{c}{Airplane} & \multicolumn{2}{c}{Chair} & \multicolumn{2}{c}{Car} \\ \cline{3-8} 
Method  & Rep. & CD           & EMD           & CD          & EMD         & CD         & EMD        \\ 
\midrule
IM-GAN~\cite{chen2019imgan} & Occupancy & 79.48 & 82.94  & 58.59 & 69.05 & 95.69 & 94.79 \\
SDF StyleGAN~\cite{zheng2022sdfstylegan} & SDF & 85.48 & 87.08 &63.25 & 67.80 & 88.34 & 88.31 \\
GET3D~\cite{gao2022get3d} & SDF & - & - & 75.26 & 72.49 & 75.26 & 72.49 \\
MeshDiffusion~\cite{liu2022meshdiffusion} & Mesh & 66.44 & 76.26 & 53.69 & 57.63 & 81.43 & 87.84  \\
LION~\cite{vahdat2022lion} & Point Cloud & 67.41 & 61.23 & 53.70 & 52.34 & 53.41 & 51.14 \\
3DQD~\cite{li2023dqd}  & TSDF & 56.29 & 54.78  & 55.61 & 52.94 & 55.75 & 52.80 \\
\midrule
\rowcolor{Gray}
Ours  & NeuSDF & \textbf{52.33} & \textbf{52.47}  & \textbf{51.95} & \textbf{52.60} & \textbf{53.06} & \textbf{51.11} \\
\bottomrule
\end{tabular}

%% file: tables/shape_comp.tex
\begin{tabular}{lllllll}
\toprule
        & \multicolumn{3}{c}{Bottom Half}   & \multicolumn{3}{c}{Octant} \\ \cline{2-7} 
Method & MMD$\downarrow$  & AMD$\downarrow$ & TMD$\uparrow$ & MMD$\downarrow$  & AMD$\downarrow$ & TMD$\uparrow$ \\ 
\midrule
PoinTr~\cite{yu2021pointr}           & 5.32 & N/A & N/A & 21.57 & N/A & N/A \\
SeedFormer~\cite{zhou2022seedformer} & 4.97 & N/A & N/A & 23.99 & N/A & N/A \\
AutoSDF~\cite{mittal2022autosdf}     & 3.51 & 8.20 & 4.66 & 5.72 & 12.79 & 8.26 \\
3DQD~\cite{li2023dqd}                & 2.93 & 6.30 & \textbf{4.78} & 4.69 & 10.93 & \textbf{9.60} \\
\midrule
\rowcolor{Gray}
Ours & \textbf{2.29} & \textbf{5.90} & 4.76  & \textbf{3.03} & \textbf{9.59} & 8.32 \\
\bottomrule
\end{tabular}

%% file: tables/image_generation.tex
\begin{tabular}{lcc}
\toprule
Method                           & CD$\downarrow$  & F-Score$\uparrow$   \\ 
\midrule
Pix2Vox~\cite{xie2019pix2vox}  & 3.00 & 0.39 \\
AutoSDF~\cite{mittal2022autosdf}  & 2.28 & 0.42 \\
SDFusion~\cite{cheng2023sdfusion} & 1.85 & 0.43 \\
\midrule
\rowcolor{Gray}
Ours                             & \textbf{0.92} & \textbf{0.61} \\ 
\bottomrule
\end{tabular}

% \begin{tabular}{lcc}
% \toprule
% Method                           & CD$\downarrow$  & F-Score$\uparrow$   \\ 
% \midrule
% Pix2Vox~\cite{xie2019pix2vox}  & 3.001 & 0.385 \\
% AutoSDF~\cite{mittal2022autosdf}  & 2.267 & 0.415 \\
% SDFusion~\cite{cheng2023sdfusion} & 1.852 & 0.432 \\
% \midrule
% Ours                             & \textbf{0.924} & \textbf{0.612} \\ 
% \bottomrule
% \end{tabular}

%% file: tables/text_generation.tex
\begin{tabular}{lcccc}
\toprule
Method                           & PMMD$\downarrow$  & CLIP-S$\uparrow$ & FPD$\downarrow$    & TMD$\uparrow$    \\ 
\midrule
Shape-IMLE~\cite{liu2022towards} & 1.68 & 31.42  & 82.34  & 0.54 \\
AutoSDF~\cite{mittal2022autosdf} & 1.96 & 31.65  & 141.87 & 1.30 \\ 
3DQD~\cite{li2023dqd}            & 1.49 & 32.11  & 59.00  & 2.80 \\
\midrule
\rowcolor{Gray}
Ours                             & \textbf{1.49} & \textbf{32.52}  & \textbf{55.01} & \textbf{3.20} \\ 
\bottomrule
\end{tabular}

% \begin{tabular}{lcccc}
% \toprule
% Method                           & PMMD$\downarrow$  & CLIP-S$\uparrow$ & FPD$\downarrow$    & TMD$\uparrow$    \\ 
% \midrule
% Shape-IMLE~\cite{liu2022towards} & 1.681 & 31.42  & 82.34  & 0.0539 \\
% AutoSDF~\cite{mittal2022autosdf} & 1.961 & 31.65  & 141.87 & 0.1302 \\ 
% 3DQD~\cite{li2023dqd}            & 1.492 & 32.11  & 59.00  & 0.2795 \\
% \midrule
% Ours                             & \textbf{1.492} & \textbf{32.52}  & \textbf{55.01} & \textbf{0.3196} \\ 
% \bottomrule
% \end{tabular}

%% file: supplementary/tables/planes.tex
\begin{tabular}{lcc}
\toprule
                & CD                       & EMD                      \\
\midrule
Channel Concat. & 4.86                     & 3.67                     \\
Roll out        & 2.05                     & 1.22                     \\ 
\midrule
Ours            & \textbf{1.76} & \textbf{0.94} \\ 
\bottomrule
\end{tabular}

%% file: supplementary/tables/sape.tex
\begin{tabular}{lcc}
\toprule
                & CD & EMD   \\
\midrule
w/o SAPE & 1.12  & 1.69                    \\
w/. SAPE & \textbf{1.04} & \textbf{1.47} \\ 
\bottomrule
\end{tabular}

%% file: supplementary/tables/ldm.tex
\begin{tabular}{lcccc}
\toprule
                    & \multicolumn{2}{c}{COV $(\%,\uparrow)$} & \multicolumn{2}{c}{1-NNA$(\%,\downarrow$)} \\ \cline{2-5} 
                    & CD         & EMD        & CD          & EMD         \\ 
\midrule
Raw Tri-plane Diff. & 28.57      & 42.86      & 64.29       & 57.14       \\
Latent Space Diff.  & \textbf{60.17}      & \textbf{54.98}      & \textbf{51.95}       & \textbf{52.60}       \\ 
\bottomrule
\end{tabular}

%% file: supplementary/tables/neusdf.tex
\begin{tabular}{lcccc}
\toprule
            & Voxel SDF Field & NeuSDF-1 & NeuSDF-2 &  NeuSDF-3\\ 
\midrule
% Rel. Speed  & $1\times$                          & $0.64\times$                           & $0.02\times$                          \\
Resolution & $64^3$ & $64^2 \times 3$ & $128^2 \times 3$ & $128^2 \times 3 \rightarrow 512$ \\
Rel. Memory ($\downarrow$) & $1\times$     & $\mathbf{0.25\times}$  & $1\times$   & $1\times$       \\ 
Recon. Error ($\downarrow$) & 36.83 & 0.61 & 0.45 & \textbf{0.38}\\
\bottomrule
\end{tabular}